\documentclass[lettersize,journal]{IEEEtran}
\usepackage{amsmath,amsfonts}
\usepackage{algorithmic}
\usepackage{algorithm}
\usepackage{array}
\usepackage[caption=false,font=normalsize,labelfont=sf,textfont=sf]{subfig}
\usepackage{textcomp}
\usepackage{stfloats}
\usepackage{url}
\usepackage{verbatim}
\usepackage{graphicx}
\usepackage{cite}
\usepackage{amssymb}
\usepackage{booktabs}
\usepackage{multirow}
\usepackage{color, colortbl}
\usepackage{subcaption}
\begin{document}

\title{Multimodal Graph Network Modeling for Human-Object Interaction Detection with PDE Graph Diffusion}

\author{\IEEEauthorblockN{Wenxuan Ji$^{1,2}$, Haichao Shi$^1$,~\IEEEmembership{Member,~IEEE,}, Xiao-Yu Zhang$^{1,*}$,~\IEEEmembership{Senior Member,~IEEE,}} \\
\IEEEauthorblockA{$^1$Institute of Information Engineering, Chinese Academy of Sciences} \\
\IEEEauthorblockA{$^2$School of Cyber Security, University of Chinese Academy of Sciences\\
Email: \{jiwenxuan, shihaichao, zhangxiaoyu\}@iie.ac.cn}
\thanks{$^*$Corresponding Author}
}

\markboth{}%
{Shell \MakeLowercase{\textit{et al.}}: A Sample Article Using IEEEtran.cls for IEEE Journals}

\IEEEpubid{}

\maketitle

\begin{abstract}
Existing GNN-based Human-Object Interaction (HOI) detection methods rely on simple MLPs to fuse instance features and propagate information. However, this mechanism is largely empirical and lack of targeted information propagation process. To address this problem, we propose Multimodal Graph Network Modeling (MGNM) for HOI detection with Partial Differential Equation (PDE) graph diffusion. Specifically, we first design a multimodal graph network framework that explicitly models the HOI detection task within a four-stage graph structure. Next, we propose a novel PDE diffusion mechanism to facilitate information propagation within this graph. This mechanism leverages multimodal features to propaganda information via a white-box PDE diffusion equation. Furthermore, we design a variational information squeezing (VIS) mechanism to further refine the multimodal features extracted from CLIP, thereby mitigating the impact of noise inherent in pretrained Vision-Language Models. Extensive experiments demonstrate that our MGNM achieves state-of-the-art performance on two widely used benchmarks: HICO-DET and V-COCO. Moreover, when integrated with a more advanced object detector, our method yields significant performance gains while maintaining an effective balance between rare and non-rare categories.
\end{abstract}


\section{Introduction}
\label{sec:intro}
Human-object interaction detection is a task that aims to localize human-object pairs and simultaneously infer the interactions between them. This task represents a more profound level of visual scene understanding than standard object detection. A detected HOI instance is represented as a triplet in the form of $\langle \text{human, action, object} \rangle$. Recently, HOI detection has garnered significant attention from the computer vision community~\cite{10227593, 8848601, wang2024interaction, Jia2024OrchestratingTS, Yang2025NoMS, Xue2025GuidingHI, 10658050}, owing to its extensive applications in downstream tasks such as action recognition~\cite{11247879, 11194256, 10812791}, image caption~\cite{10097833, 10844064, 9410374}, and visual scene understanding~\cite{ma2024scene, 11264347}.

Similar to object detectors, HOI detectors can be categorized into two primary types: one-stage and two-stage methods. Recent one-stage methods usually employ the Transformer architecture~\cite{Vaswani2017AttentionIA}. Specifically, these methods often adapt DETR~\cite{Carion2020EndtoEndOD} to detect human-object pairs and predict the interactions in parallel via Transformer decoders~\cite{Zhou2022HumanObjectID, Jia2024ContextHOISC}. However, compared with two-stage methods, one-stage methods cannot easily leverage the features from object detectors, potentially limiting their performance. Two-stage methods typically follow a distinct training paradigm, which generally involves fine-tuning a pretrained object detector and subsequently freezing its parameters~\cite{zhang2022upt, zhang2023pvic, park2023viplo}. This paradigm allows the interaction predictor to focus solely on interaction prediction, thereby improving training efficiency. 

\begin{figure}[t]
  \centering
  \includegraphics[width=1.0\linewidth]{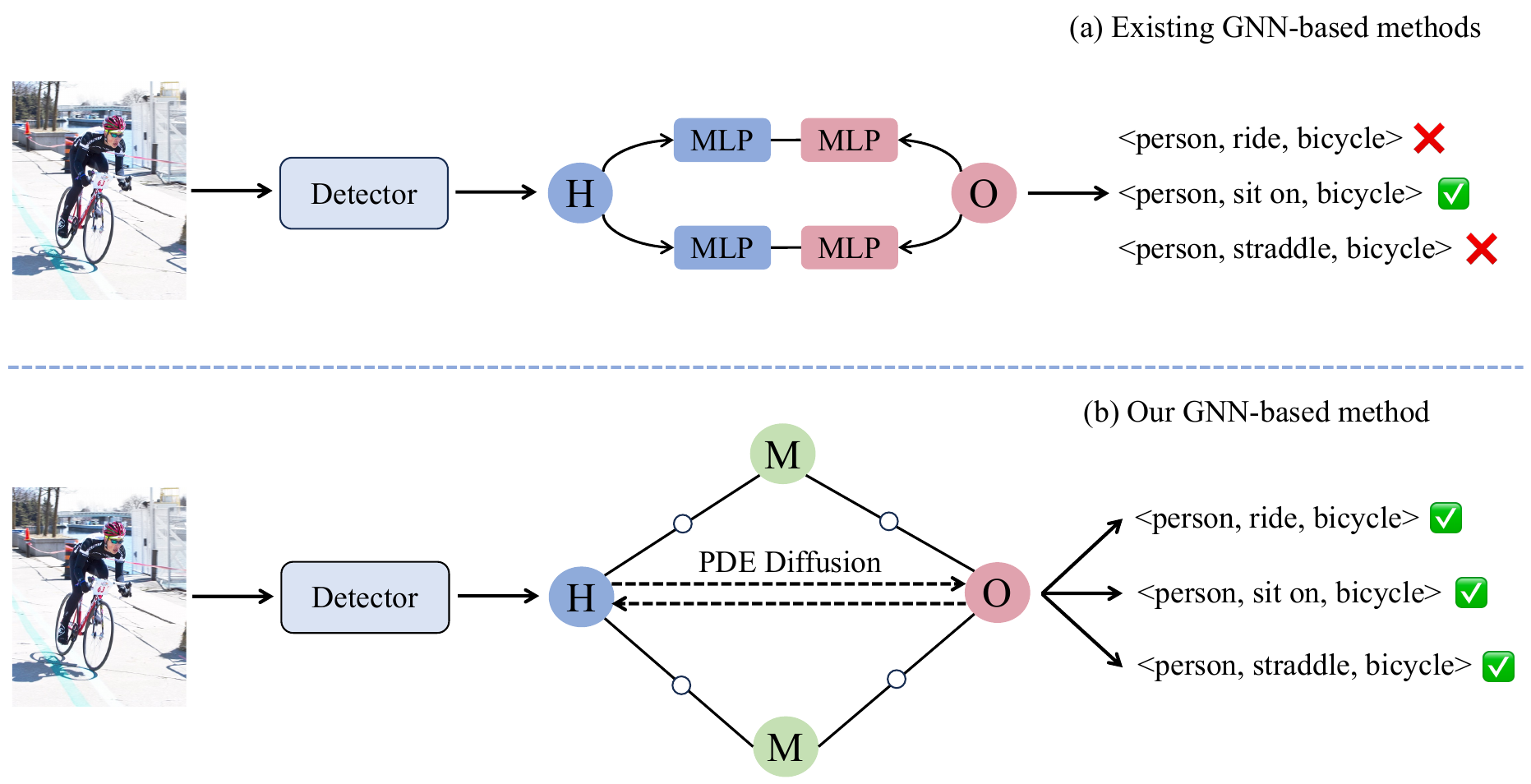}
  \caption{\textbf{Comparison between existing GNN-based methods and our proposed method.} Previous GNN-based methods usually employ a black-box MLP fusion strategy to extract messages. In contrast, our proposed approach utilizes a white-box PDE diffusion mechanism to propagate messages, while incorporating multimodal features to provide additional contextual cues. Here, H, O, and M denote human, object, and multimodal features, respectively.}
  \label{fig:compare}
\end{figure}

As illustrated in Fig.~\ref{fig:compare}(a), recent GNN-based two-stage methods typically employ simple MLPs to split, transform, and reassemble the instance features for message generation. However, this black-box fusion mechanism lacks interpretability and targeted information propagation process in GNNs. Moreover, their exclusive focus on instance features leads to a lack of contextual semantics, thereby hindering further performance improvements. Our motivation about GNN-based HOI methods is: interaction prediction comes from information propagation. Consequently, we propose a multimodal graph network framework with PDE graph diffusion, which provides fine-grained, discriminative cues for interaction prediction.

As illustrated in Fig.~\ref{fig:compare}(b), drawing inspiration from the physical heat equation, our method adopts a PDE diffusion mechanism to explicitly model the information propagation process within the graph. Meanwhile, as observed in prior studies~\cite{zhang2021scg, park2023viplo}, relying solely on the visual features of human and object instances often fails to capture fine-grained contextual cues. Motivated by the success of Vision-Language Models (VLMs)~\cite{Liao2022GENVLKTSA, Lei2024EZHOIVA, Guo2024UnseenNM, Ning2023HOICLIPEK}, we leverage CLIP~\cite{Radford2021LearningTV} to capture rich multimodal features. Furthermore, to eliminate scene noise in these multimodal features, we propose the variational
information squeezing (VIS) mechanism, which extracts the most valuable contextual cues by optimizing the variational upper bound. These multimodal features break the limitations of instance features
and provide more fine-grained contextual cues for our framework. Thus, we introduce the multimodal graph network framework. It leverages both multimodal features and PDE diffusion to enhance the information propagation between human-object pairs, thereby enriching the graph network with more fine-grained semantic information.

In light of the preceding analysis, we propose \textbf{M}ultimodal \textbf{G}raph \textbf{N}etwork \textbf{M}odeling (MGNM), a novel and effective two-stage framework for HOI detection. Our MGNM achieves state-of-the-art (SOTA) performance on two widely used benchmarks: HICO-DET~\cite{chao2018learning} and V-COCO~\cite{gupta2015visual}. Furthermore, benefiting from the rich prior knowledge within the multimodal features, our method not only achieves substantial performance improvements but also effectively mitigates the performance disparity between rare and non-rare categories, a persistent challenge for previous methods~\cite{zhang2021scg, zhang2022upt, liao2020ppdm, 10315071}. Notably, when integrated with a more advanced object detector, our method demonstrates a significant performance gain and preserves the effective balance between rare and non-rare categories. The effectiveness and efficiency of our framework is validated through extensive ablation studies. The main contributions of this work are summarized as follows:
\begin{itemize}
    \item {We propose a novel multimodal graph network framework that leverages multimodal features extracted from CLIP, thus overcoming the limitations of previous GNN-based HOI methods that rely solely on instance features.}
    
    \item {Inspired by the physical heat equation, we propose a novel PDE diffusion mechanism that employs a white-box PDE formulation to enhance the information propagation within GNN.}

    \item {We propose a novel variational information squeezing mechanism, designed to capture more compact and task-relevant multimodal features for HOI detection.}
    
    \item {Extensive experiments on HICO-DET and V-COCO benchmarks demonstrate that our method not only achieves SOTA performance but also maintains an effective balance between rare and non-rare categories.}
\end{itemize}

The remainder of this paper is organized as follows. In section~\ref{sec:related}, we review related work in the field of HOI detection. Section~\ref{sec:method} details our proposed framework in details. Next, in section~\ref{sec:exper}, we compare our proposed MGNM with other SOTA methods and analyze its effectiveness and efficiency. Subsequently, section~\ref{sec:qua} presents qualitative results and discusses the limitations of our approach. Finally, section~\ref{sec:con} concludes the paper.

\section{Related Work} 
\label{sec:related}
\subsection{One-stage Methods}
One-stage methods are characterized by their parallel detection of human and object instances and prediction of their corresponding interactions. Early approaches typically relied on heuristics such as human keypoints~\cite{liao2020ppdm, Zhong2021GlanceAG} or predefined union boxes~\cite{kim2020uniondet, Gkioxari2017DetectingAR}. More recently, Transformer-based architectures have become the dominant paradigm for one-stage HOI detection. For instance, Chen et al.~\cite{chen_2021_asnet} employ two separate Transformer decoders to predict human and object instances and their interactions in parallel. Similarly, Yang et al.~\cite{yang2023boosting} leveraged VLMs to extract fine-grained semantic features, which were then processed by two sub-decoders to simultaneously predict HOI triplets. Nevertheless, one-stage methods are often encumbered by significant memory usage and slow convergence speed.

\subsection{Two-stage Methods}
Two-stage methods first employ an off-the-shelf object detector to identify human and object instances, and subsequently predict interactions between the matched human-object pairs. Recent two-stage methods often use Transformer to extract features or enhance attention mechanisms~\cite{zhang2022upt, zhang2023pvic}. Another line of work~\cite{Guo2024UnseenNM, 11092421, lei2024efficient, NEURIPS2024_2a54def4, 11249431} has incorporated VLMs for feature refinement, and subsequently adopted Transformer decoders to fuse the features. For instance, Chen et al.~\cite{11249431} first generated the affordance-scene knowledge graph, and then input these prompts to CLIP to capture fine-grained textual features. Lei et al.~\cite{lei2024efficient} utilized the Large Language Models (LLMs) to generate fine-grained descriptions for human-object pairs. Then, they also adopted the CLIP to capture fine-grained visual and language features for final interaction prediction. In comparison, our work is situated within a growing line of GNN-based HOI detectors~\cite{Qi2018LearningHI, Gao2020DRGDR, 9489275, zhang2021scg, park2023viplo}. In this line, SCG~\cite{zhang2021scg} utilized a GNN to facilitate information exchange between human and object nodes, whereas ViPLO~\cite{park2023viplo} incorporated features from the CLIP image encoder and ViTPose-L~\cite{xu2022vitpose} to further enrich human and object instance representations. Nevertheless, a key limitation of these methods is that they typically employ a black-box MLP fusion mechanism to propagate the information within GNN. Moreover, they primarily rely on visual instance features, overlooking the contextual semantics provided by multimodal features, thereby limiting their further performance improvement.

\section{Method}
\label{sec:method}
\begin{figure*}[t]
  \centering
  \includegraphics[width=1.0\linewidth]{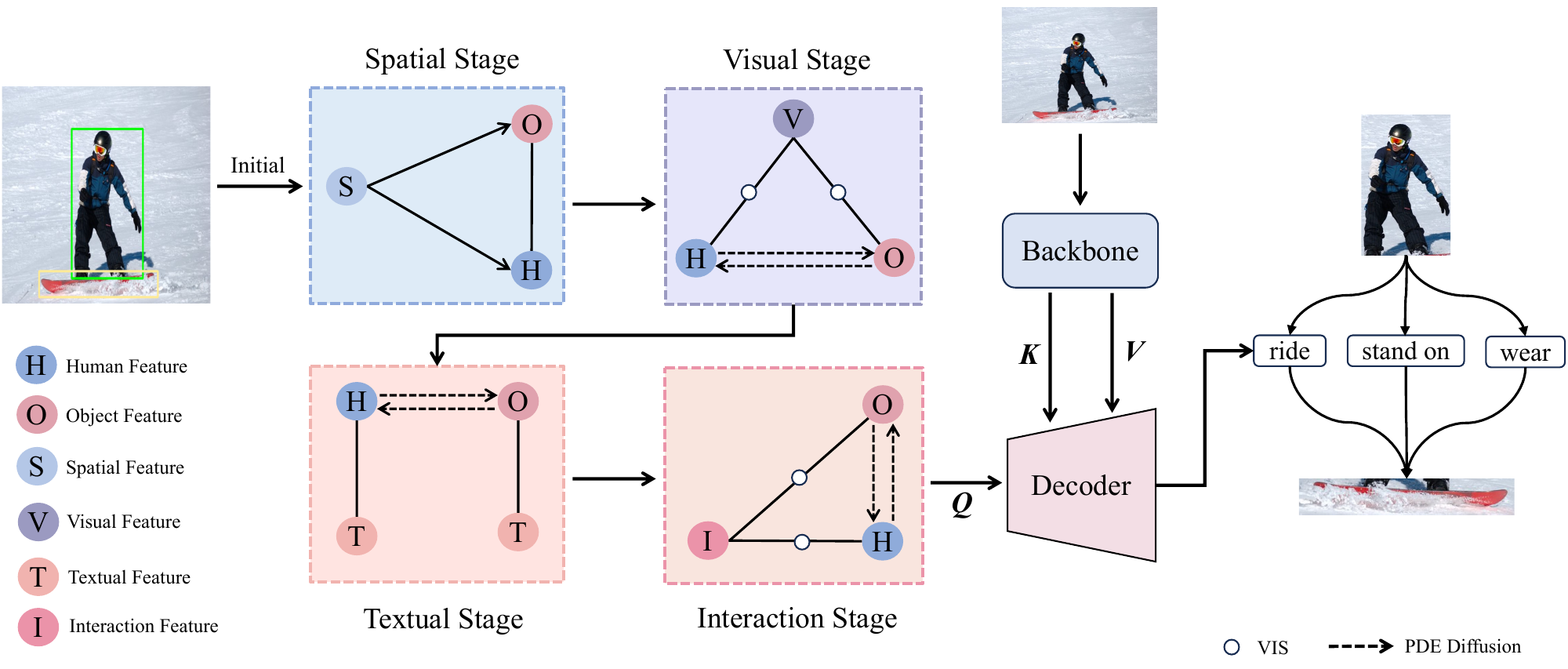}
  \caption{\textbf{Overview of our MGNM framework}. The core of our proposed MGNM framework is a four-stage multimodal graph network. (1) Spatial Stage: For each candidate pair, low-level spatial features derived from their bounding boxes are utilized to initialize pairwise representations. Subsequently, these spatial features are employed to construct an adjacency matrix that regulates feature interactions in the subsequent stages. (2) Visual Stage: High-level visual semantic features are extracted using the CLIP image encoder. Facilitated by the VIS and PDE diffusion mechanisms, these visual cues further enrich information propagation between human-object pairs. (3) Textual Stage: The CLIP text encoder is adopted to obtain semantic cues for the corresponding human and object instances. (4) Interaction Stage: In conjunction with the image-level prompt, we again employ the CLIP text encoder to captures high-level interaction features between human-object pairs, thereby facilitating more effective relational reasoning within the graph structure. Finally, the refined pairwise representations are fed as queries into the Transformer decoder to predict the final HOI triplets.}
  \label{fig:method}
\end{figure*}

This section details the proposed method. We first elaborate on the architecture of the four-stage multimodal graph network framework. Subsequently, we present the details of our proposed PDE diffusion and VIS mechanisms. Finally, we introduce the training and inference process of our approach. An overview of our method is illustrated in Fig.~\ref{fig:method}. 

In this paper, we assume a given image contains $n$ detected objects (including $m$ humans). With a standard one-to-one matching strategy, we obtain $m \times n$ candidate human-object pairs, which can be seen as the adjacency matrix in GNN. Here, we denote $\mathbf{N} \in \mathbb{R}^{n \times d}$ as the human and object instances set, which serves as the node representations in GNN. The corresponding human and object sets within the human-object pairs are represented as $\mathbf{H} \in \mathbb{R}^{(m \times n) \times d}$ and $\mathbf{O} \in \mathbb{R}^{(m \times n) \times d}$. $d$ is the dimension of the nodes. Additionally, the features of human-object pairs are denoted as $\mathbf{E} \in \mathbb{R}^{(m \times n) \times 2d}$.

\subsection{Overall Framework}
\label{sec:method:overall}
\begin{algorithm}[tb]
\caption{Computation Process of Interaction Prediction}
\label{alg}
\textbf{Input}: $\mathbf{N}, \mathbf{H}, \mathbf{O}$, Image\\
\textbf{Output}: $\text{Decoder}(\mathcal{Q}, \mathcal{K}, \mathcal{V})$
\begin{algorithmic}[1] 
\STATE $\mathbf{E} = \text{MLP}\left(\text{Linear}(\mathbf{H} \oplus \mathbf{O}) \oplus \text{Linear}(\mathbf{S})\right)$
\FOR{$i=1:T$}
\STATE $\mathbf{A} = \text{Linear}(\text{MBF}(\mathbf{H} \oplus \mathbf{O}, \mathbf{S}))$
\STATE $\mathbf{N} = \text{LN}(\text{PDEDiff}_{\mathcal{V}}(\mathbf{N}, \mathbf{A}, \text{VIS}(\mathbf{V})))$
\STATE $\mathbf{N} = \text{LN}(\text{PDEDiff}_{\mathcal{T}}(\mathbf{N}, \mathbf{A}, \mathbf{T}))$
\STATE $\mathbf{N} = \text{LN}(\text{PDEDiff}_{\mathcal{I}}(\mathbf{N}, \mathbf{A}, \text{VIS}(\mathbf{I})))$
\STATE $\mathbf{E} = \text{LN}(\mathbf{E}+\mathbf{H} \oplus \mathbf{O})$
\ENDFOR
\STATE $\mathcal{Q}=\mathbf{E}$; $\mathcal{K,V}=\text{Backbone}(\text{Image})$
\STATE \textbf{return} $\text{Decoder}(\mathcal{Q}, \mathcal{K}, \mathcal{V})$
\end{algorithmic}
\end{algorithm}

Our method initially employs an off-the-shelf object detector to extract bounding boxes and appearance features. Subsequently, as shown in Fig.~\ref{fig:method} and Algorithm~\ref{alg} (Lines 1-8), our multimodal graph network is structured into four sequential stages: spatial, visual, textual, and interaction. This multi-stage architecture leverages both low-level and high-level visual and language features to enhance the interactive modeling of human-object pairs. The details of each stage are elaborated below.

\paragraph{Spatial Stage}
The spatial stage is designed to initialize the human-object pair representations and exploit geometric relationships to enhance the modeling of interactions between them. Specifically, the spatial features comprise bounding box coordinates, box areas, intersection over union (IoU), and other related geometric metrics. Further details regarding these spatial features can be found in Appendix A. These features encode the fundamental geometric relationships of each human-object pair and serve as a source of low-level visual information. Moreover, since these are handcrafted features, their computation incurs minimal overhead for the model. Upon extracting the spatial features, we utilize a simple MLP fusion strategy to integrate them with the base human and object features, thereby generating the initial representations for each human-object pair:
\begin{align}
    \mathbf{E} &= \text{MLP}\left(\text{Linear}(\mathbf{H} \oplus \mathbf{O}) \oplus \text{Linear}(\mathbf{S})\right),
\end{align}
where the linear layer is used to align the dimensions. $\oplus$ means the concatenation operation and $\mathbf{S}$ means the spatial features.

Furthermore, leveraging the spatial features alongside the corresponding human and object features, the adjacency matrix $\mathbf{A} \in \mathbb{R}^{m \times n}$ for the multimodal graph network is constructed via the MBF fusion mechanism~\cite{zhang2021scg}:
\begin{align}
    \mathbf{A} &= \text{Linear}\left(\text{MBF}\left(\mathbf{H} \oplus \mathbf{O}, \mathbf{S}\right)\right),
\end{align}

\paragraph{Visual Stage}
The visual stage leverages CLIP to integrate high-level visual features. Specifically, we employ the CLIP image encoder to extract high-level visual semantic features from the input image. Subsequently, we apply the VIS mechanism to refine these extracted visual features. These refined features are then used to propagate the high-level visual semantic cues to human and object nodes via the PDE diffusion mechanism, formulated as follows:
\begin{align}
    \mathbf{N} &= \text{LN}(\text{PDEDiff}_{\mathcal{V}}(\mathbf{N}, \mathbf{A}, \text{VIS}(\mathbf{V}))),
\end{align}
where LN means the layer normalization. $\text{PDEDiff}_{\mathcal{V}}$ denotes the PDE diffusion mechanism responsible for propagating visual information to the human and object nodes. $\mathbf{V}$ represents the visual features from CLIP. 

\paragraph{Textual Stage}
The textual stage is dedicated to generating textual embeddings for each node, thereby providing a source of low-level language features. As illustrated in Fig.~\ref{fig:method}, we adopt a simple object-centric prompt template ``a photo of $\langle \text{object} \rangle$", to focus on the target object category. Here, the $\langle \text{object} \rangle$ corresponds to the object label derived from the detection results. Given that these prompts are simple and clear, we do not apply the VIS mechanism in this stage. This process is formulated as follows:
\begin{align}
    \mathbf{N} &= \text{LN}(\text{PDEDiff}_{\mathcal{T}}(\mathbf{N}, \mathbf{A}, \mathbf{T})),
\end{align}
where $\text{PDEDiff}_{\mathcal{T}}$ denotes the PDE diffusion mechanism responsible for propagating textual information to the human and object nodes. $\mathbf{T}$ represents the textual embeddings extracted via the CLIP text encoder.

\paragraph{Interaction Stage}
To incorporate high-level interaction semantic cues, we adopt an interaction-centric prompt: ``a scene containing people and $\langle \text{objects} \rangle$ with potential interactions". Here, $\langle \text{objects} \rangle$ represents the object labels derived from the detection results. Notably, we set a confidence threshold of 0.05 to filter out low-confidence object predictions. We opt for an image-level prompt design as it is more efficient for HOI detection. Although HOI-instance-level prompts might yield greater performance improvements, they will introduce substantial computational overhead, significantly slowing down inference speed. Leveraging this designed interaction-centric prompt, we once again employ the CLIP text encoder. Specifically, it is utilized to generate interaction-centric semantic features, which constitute high-level language features. This process is formulated as follows:
\begin{align}
    \mathbf{N} &= \text{LN}(\text{PDEDiff}_{\mathcal{I}}(\mathbf{N}, \mathbf{A}, \text{VIS}(\mathbf{I}))),
\end{align}
where $\mathbf{I}$ denotes the interaction features of the human-object pairs. Subsequently, we concatenate the node representations to update the initial human-object pair representations:
\begin{align}
    \mathbf{E} &= \text{LN}(\mathbf{E}+\mathbf{H} \oplus \mathbf{O}),
\end{align}
Finally, as described in Algorithm~\ref{alg} (Line 9), the human-object pair representations are utilized as queries within the decoder layer to predict the final interactions.

\subsection{PDE Graph Diffusion}
\begin{figure}[t]
  \centering
  \includegraphics[width=1.0\linewidth]{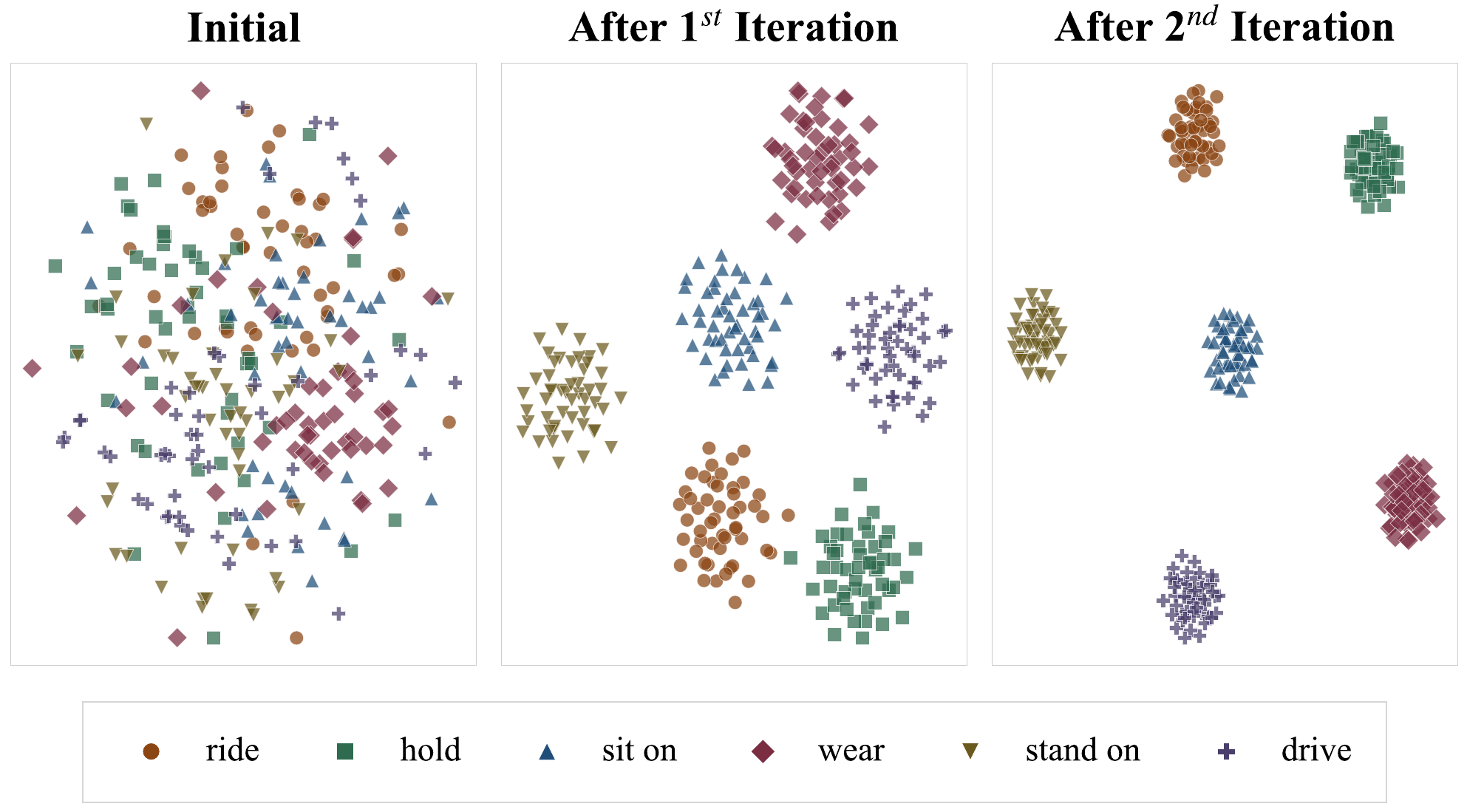}
  \caption{T-SNE visualization of our PDE diffusion mechanism.}
  \label{fig:tsne}
\end{figure}

To explicitly model the information propagation process within the graph, we propose a novel PDE graph diffusion mechanism inspired by the physical heat equation. Given the bipartite graph characteristics inherent to HOI tasks, we formulate the energy function as follows:
\begin{align}
    \mathcal{E}(\mathbf{N})&=\underbrace{\frac{1}{2} \sum_{i, j}\left(\mathbf{A}_{N^h \rightarrow N}\right)_{i j}\left\|\mathbf{N}_i-\mathbf{N}^h_j\right\|^2}_{\text {Action Potential } \mathcal{E}_{\text {act}}} \notag \\
    &+\underbrace{\frac{1}{2} \sum_{j, i}\left(\mathbf{A}_{N \rightarrow N^h}\right)_{j i}\left\|\mathbf{N}^h_j-\mathbf{N}_i\right\|^2}_{\text {Affordance Potential } \mathcal{E}_{\text {afford }}} \notag \\
    &+\underbrace{\frac{\lambda}{2}\|\mathbf{N}-\tilde{\mathbf{M}}\|}_{\text{Variational Fidelity } \mathcal{E}_{\text {vlm}}},
\end{align}
where $\mathbf{N}^h \in \mathbb{R}^{m \times d}$ denotes the human nodes in $\mathbf{N}$. Unlike the standard heat equation, our proposed PDE diffusion mechanism incorporates a human action potential $\mathcal{E}_{\text {act}}$ and an object affordance potential $\mathcal{E}_{\text {afford}}$. These two adjacency matrices are derived from the bipartite adjacency matrix $\mathbf{A}$ via a softmax operation. The variational fidelity $\mathbf{\mathcal{E}_{\text {vlm}}}$ is used to constrain the human and object instances using the refined multimodal features, thereby infusing the instances with contextual semantics from an external VLM. 

Grounded in the principles of the heat equation, the information propagation process is equivalent to minimizing this energy function. Specifically, we map the discrete graph network onto a negative gradient flow over continuous time $t$. This continuous diffusion process is formulated as follows:
\begin{align}
\frac{\partial \mathbf{N}^h}{\partial t}=-\frac{\partial \mathcal{E}}{\partial \mathbf{N}^h}&=\underbrace{\left(\mathbf{A}_{N^h \rightarrow N} \mathbf{N}^h-\mathbf{N}^h\right)}_{\text {Action }} \notag \\
&+\underbrace{\left(\mathbf{A}_{N \rightarrow N^h} \mathbf{N}-\mathbf{N}^h\right)}_{\text {Affordance }}+\underbrace{\lambda\left(\tilde{\mathbf{M}}-\mathbf{N}^h\right)}_{\text {Fidelity }}, \label{eq:8} \\
\frac{\partial \mathbf{N}}{\partial t}=-\frac{\partial \mathcal{E}}{\partial \mathbf{N}}
&= \underbrace{\left(\mathbf{A}_{N^h \rightarrow N} \mathbf{N}^h - \mathbf{N}\right)}_{\text{Action }}+ \underbrace{\lambda\left(\tilde{\mathbf{M}} - \mathbf{N}\right)}_{\text{Fidelity }}. \label{eq:9}
\end{align}
During this PDE diffusion process, human nodes simultaneously receive both action and affordance potential information, as humans serve as the subject and object simultaneously in HOI detection. The fidelity term serves as an external information source for the PDE, providing the contextual semantics for the nodes.

Finally, we apply the forward-euler discretization method to numerically solve the aforementioned PDE diffusion equations as follows:
\begin{align}
\mathbf{N}^{l+1}&=\mathbf{N}^{l}+\tau \cdot \Delta \mathbf{N}_L^{l}+\gamma \cdot \Delta \mathbf{N}_S^{l},
\end{align}
where $l$ is the iteration steps. $\tau$ and $\gamma$ are learnable parameters. $\Delta \mathbf{N}_L^{l}$ and $\Delta \mathbf{N}_S^{l}$ denote the laplacian flow and the fidelity flow, respectively. For human nodes, $\Delta \mathbf{N}_L^{l}$ is the superposition of action and affordance potentials as defined in Eq.~\ref{eq:8}, whereas for object nodes, it is derived solely from the action potential in Eq.~\ref{eq:9}. The fidelity flow serves to preserve contextual cues extracted from multimodal features throughout the PDE diffusion process. As illustrated in Fig.~\ref{fig:tsne}, our PDE diffusion mechanism rapidly achieves the optimal performance and significantly improves feature distinction across different actions after 2 iterations. 

\subsection{Variational Information Squeezing}
Although the multimodal features from CLIP have been demonstrated to be effective~\cite{Lei2024EZHOIVA, ting2023hoi}, a domain gap persists between these representations and the HOI detection task. Because CLIP is pretrained on image-level contrastive learning tasks, its features inevitably contain noise that is irrelevant to HOI detection. To address this limitation, we design a variational information squeezing mechanism to filter out irrelevant noise from the multimodal features. The optimization objective can be formulated as follows:
\begin{align}
    \max _\theta\left(I(\tilde{\mathbf{M}} ; Y)-\beta \cdot I(\tilde{\mathbf{M}} ; \mathbf{M})\right),
\end{align}
where $I(\tilde{\mathbf{M}} ; Y)$ represents the mutual information between the refined features and the prediction results. $I(\tilde{\mathbf{M}} ; \mathbf{M})$ denotes the mutual information between the refined features and the original multimodal features. $\theta$ denotes the set of learnable parameters within the network. Due to the intractability of directly computing mutual information, we employ a variational inference approximation~\cite{Kingma2013AutoEncodingVB}. The process of deriving these refined features is formulated as follows:
\begin{align}
    q_\theta(\tilde{\mathbf{M}} \mid \mathbf{M})&=\mathcal{N}\left(\mu_{\mathbf{M}}, \operatorname{diag}\left(\sigma_{\mathbf{M}}^2\right)\right), \\
    \left[\mu_{\mathbf{M}}, \log \sigma_{\mathbf{M}}^2\right]&=\mathrm{MLP}_\theta(\mathbf{M}).
\end{align}
Subsequently, the upper bound of the mutual information $I(\tilde{\mathbf{M}}; \mathbf{M})$ is constrained by the KL divergence between the variational posterior distribution $q_{\theta}$ and the prior distribution $p(\tilde{\mathbf{M}})$ formulated as follows:
\begin{align}
    I(\tilde{\mathbf{M}} ; \mathbf{M}) \leq \mathbb{E}_{\mathbf{M}}\left[D_{K L}\left(q_\theta(\tilde{\mathbf{M}} \mid \mathbf{M}) \| p(\tilde{\mathbf{M}})\right)\right].
\end{align}
Following established practices in variational inference~\cite{Kingma2013AutoEncodingVB}, we assume an isotropic standard Gaussian prior, namely $p(\tilde{\mathbf{M}})=\mathcal{N}(0, \mathbf{I})$. Consequently, the KL divergence can be analytically computed as follows:
\begin{align}
    \mathcal{L}_{\text{VIS}}&=\beta \cdot D_{K L}\left(\mathcal{N}\left(\mu_{\mathbf{M}}, \sigma_{\mathbf{M}}^2\right) \| \mathcal{N}(0, \mathbf{I})\right) \notag \\
    &=-\frac{\beta}{2} \sum_{k=1}^d\left(1+\log \left(\sigma_{\mathbf{M}, k}^2\right)-\mu_{\mathbf{M}, k}^2-\sigma_{\mathbf{M}, k}^2\right),
\end{align}
where $\mathcal{L}_{\text{VIS}}$ serves as a component of the overall loss function and $\beta$ is a control parameter. By employing the reparameterization trick, the refined multimodal features can be sampled during training as follows:
\begin{align}
    \tilde{\mathbf{M}}=\mu_{\mathbf{M}}+\sigma_{\mathbf{M}} \odot \epsilon \quad \text { where } \epsilon \sim \mathcal{N}(0, \mathbf{I}).
\end{align}
During inference, $\tilde{\mathbf{M}}=\mu_{\mathbf{M}}$. Consequently, by incorporating this VIS mechanism, the resulting multimodal features are rendered purer and highly discriminative.

\subsection{Training and Inference}
Following previous works~\cite{zhang2023pvic, kim2025locality}, we incorporate the confidence scores from the detection results into the final scores for each human-object pair. In particular, given the interaction logits $s$, the final score $\widehat{s}$ is computed as follows:
\begin{equation}
    \widehat{s} = (s_h \cdot s_o)^\alpha \cdot \sigma (s),
\end{equation}
where $s_h$ and $s_o$ represent the confidence scores of the corresponding human and object within the pair and $\sigma$ denotes the sigmoid function. As HOI detection is a multi-label task, we apply sigmoid function rather than softmax function. The hyperparameter $\alpha$ is used to suppress overconfident objects during inference. Following prior works~\cite{zhang2023pvic, kim2025locality}, we empirically set $\alpha=1$ during training and $\alpha=2.5$ during inference.

To alleviate the imbalance problem of positive and negative samples in the HOI detection task, we employ the focal loss with logits~\cite{Lin2017FocalLF} as the loss function during the training process:
\begin{equation}
    \mathcal{L}_\text{HOI} = \text{FL}(\text{Logit}(\widehat{s}), s_{GT}),
\end{equation}
where FL denotes the focal loss function, Logit represents the standard logit transformation, and $s_{GT}$ denotes the ground-truth labels. With the aforementioned loss in VIS mechanism, the complete loss function is as follows:
\begin{equation}
    \mathcal{L}=\mathcal{L}_\text{HOI}+\beta * \mathcal{L}_\text{VIS},
\end{equation}
where $\beta$ is a control parameter and set as 0.01.

\section{Experiments}
\label{sec:exper}

\begin{table*}[t]
\begin{center}
\caption{Performance comparison in terms of mAP (\%) on the HICO-DET and V-COCO dataset. In each evaluation metric, the best result is marked in bold and the second-best result is underlined. In this paper, R50, R101, and R101-DC denotes ResNet50, ResNet101, and ResNet101 with dilated convolution, respectively. SD means Stable Diffusion model~\cite{Rombach2021HighResolutionIS}.}
\label{tab:compare-all}
\begin{tabular}{llcccccccccc}
\hline
& & \multicolumn{7}{c}{\textbf{HICO-DET}} && \multicolumn{2}{c}{\textbf{V-COCO}} \\
\multirow{2}{*}{Method} & \multirow{2}{*}{Backbone} & \multicolumn{3}{c}{Default} && \multicolumn{3}{c}{Known Object} \\
\cmidrule{3-5} \cmidrule{7-9} \cmidrule{11-12}
&& Full & Rare & Non-Rare && Full & Rare & Non-Rare && $\text{AP}^{S1}_{role}$ & $\text{AP}^{S2}_{role}$  \\
\hline
\multicolumn{5}{l}{\emph{One-stage methods}} \\
UnionDet~\cite{kim2020uniondet} & R50 & 17.58 & 11.72 & 19.33 && 19.76 & 14.68 & 21.27 && 47.5 & 56.2 \\
PPDM~\cite{liao2020ppdm} & Hourglass & 21.73 & 13.78 & 24.10 && 24.58 & 16.65 & 26.84 && - & - \\
AS-Net~\cite{chen_2021_asnet} & R50 & 28.87 & 24.25 & 30.25 && 31.74 & 27.07 & 33.14 && 53.9 & -\\
FGAHOI~\cite{10315071} & Swin-T & 29.94 & 22.24 & 32.24 && 32.48 & 24.16 & 34.97 && 60.5 & 61.2 \\
CDT~\cite{10242152} & R50 & 30.48 & 25.48 & 32.37 && - & - & - && 61.4 & 65.4 \\
DOQ~\cite{qu2022distillation} & R50+CLIP & 33.28 & 29.19 & 34.50 && - & - & - && 63.5 & - \\
DiffHOI~\cite{yang2023boosting} & R50+CLIP+SD & 34.41 & 31.07 & 35.40 && 37.31 & 34.56 & 38.14 && 61.1 & 63.5 \\
MP-HOI~\cite{Yang2024OpenWorldHI} & R50+CLIP+SD & 36.50 & 35.48 & 36.80 && - & - & - && 66.2 & 67.6 \\
FGAHOI-L~\cite{10315071} & Swin-L & 37.18 & 30.71 & 39.11 && 38.93 & 31.93 & 41.02 && - & - \\
DiffHOI-L~\cite{yang2023boosting} & Swin-L+CLIP+SD & 40.63 & 38.10 & 41.38 && 43.14 & 40.24 & 44.01 && 65.7 & 68.2 \\
MP-HOI-L~\cite{Yang2024OpenWorldHI} & Swin-L+CLIP+SD & 44.53 & 44.48 & 44.55 && - & - & - && - & - \\
\hline
\multicolumn{5}{l}{\emph{Two-stage methods}} \\
SCG~\cite{zhang2021scg} & R50 & 29.26 & 24.61 & 30.65 && 32.87 & 27.89 & 34.35 && 54.2 & 60.9 \\
PPDM++~\cite{10496247} & Swin-B & 30.10 & 23.73 & 32.00 && 31.80 & 24.93 & 33.85 && - & - \\
UPT~\cite{zhang2022upt} & R101 & 32.31 & 28.55 & 33.44 && 35.65 & 31.60 & 36.86 && 60.7 & 66.2 \\
ILCN~\cite{lu2025intra} & Swin-T & 33.80 & 29.83 & 34.99 && - & - & - && 63.4 & 65.3 \\
ADA-CM~\cite{ting2023hoi} & R50+CLIP & 33.80 & 31.72 & 34.42 && - & - & - && 56.1 & 61.5 \\
PViC~\cite{zhang2023pvic} & R50 & 34.69 & 32.14 & 35.45 && 38.14 & 35.38 & 38.97 && 62.8 & 67.8 \\
HOIGen~\cite{Guo2024UnseenNM} & R50+CLIP & 34.84 & 34.52 & 34.94 && - & - & - && - & - \\
ViPLO~\cite{park2023viplo} & R101-DC+CLIP+ViTPose & 34.95 & 33.83 & 35.28 && 38.15 & 36.77 & 38.56 && 60.9 & 66.6 \\
TKCE~\cite{10889833} & R50+CLIP & 35.11 & 34.35 & 35.34 && - & - & - && - & - \\
HOLa~\cite{lei2025hola} & R50+CLIP+GPT & 35.41 & 34.35 & 35.73 && - & - & - && - & - \\
LAIN~\cite{kim2025locality} & R50+CLIP & 36.02 & 35.70 & 36.11 && - & - & - && 63.4 & 65.3 \\
VRDiff~\cite{Cao_2025_ICCV} & R50+CLIP+SD & 36.77 & 35.66 & 37.11 && 39.98 & 38.69 & 40.36 && - & - \\
RLIPv2~\cite{Yuan2023RLIPv2FS} & Swin-T+BLIP & 38.60 & 33.66 & 40.07 && - & - & - && 68.8 & 70.8 \\
HORP~\cite{11092421} & R50+CLIP+Gaze & 38.61 & 36.14 & 39.34 && 40.98 & 38.25 & 41.79 && 65.6 & 68.3 \\
PViC-L~\cite{zhang2023pvic} & Swin-L & 44.32 & 44.61 & 44.24 && 47.81 & 48.38 & 47.64 && 64.1 & 70.2 \\
Pose-aware~\cite{wu2024exploring} & Swin-L+ViTPose & 46.01 & 46.74 & 45.80 && 49.50 & 50.59 & 49.18 && 63.0 & 68.7 \\
HORP-L~\cite{11092421} & Swin-L+CLIP+Gaze & \underline{47.53} & \underline{46.81} & \underline{47.74} && \underline{51.24} & \underline{50.78} & \underline{51.38} && 68.9 & \underline{71.1} \\
\hline
MGNM & R50+CLIP & 39.43 & 38.69 & 39.65 && 42.47 & 41.85 & 42.66 && \underline{68.9} & 70.6 \\
$\text{MGNM}_{L}$ & Swin-L+CLIP & \textbf{50.16} & \textbf{51.65} & \textbf{49.72} && \textbf{53.95} & \textbf{54.91} & \textbf{53.66} && \textbf{71.2} & \textbf{74.5} \\
\hline
\end{tabular}
\end{center}
\end{table*}

This section presents a series of experiments designed to evaluate the effectiveness and efficiency of our proposed method. First, we briefly introduce the experimental settings. Next, we present a comprehensive comparison of our method against the SOTA methods. Then, we conduct the ablation study of our method. All of its experiments are conducted on the HICO-DET dataset under Deafult setting. Subsequently,  we analyze the efficiency of our proposed method and compare it with other related methods. Finally, we analyze the Rare-Non-rare bias in the HOI detection field and provide some promising suggestions. 

\subsection{Experimental Settings}
\paragraph{Datasets and Evaluation Metrics}
Our experiments are conducted on two widely used public benchmarks: HICO-DET and V-COCO. The HICO-DET benchmark comprises 38,118 training and 9,658 test images. It spans 80 object categories (identical to MS COCO) and 117 action classes, yielding a total of 600 distinct HOI instance types. In contrast, V-COCO is a smaller subset of MS COCO, consisting of 2,533 training, 2,867 validation, and 4,946 test images, covering 80 object classes and 24 action classes. Following standard protocols, a predicted HOI triplet is considered a true positive if and only if the IoU with the ground truth for both human and object boxes exceeds 0.5, and the interaction category is predicted correctly. For the HICO-DET dataset, we report performance under two standard evaluation settings: Default and Known Object. Each setting has three different subsets: (1) all 600 HOI classes (Full), (2) 138 HOI classes with less than 10 training instances (Rare), and (3) 462 HOI classes with 10 or more training instances (Non-Rare). For the V-COCO dataset, we evaluate performance based on two scenarios: Scenario 1, which requires the correct prediction of occluded object bounding boxes, and Scenario 2, which does not.

Following the standard evaluation protocols in~\cite{zhang2023pvic, park2023viplo}, we report the mean average precision (mAP) for evaluation. Given its larger scale and closer alignment with real-world data distributions, our primary experimental focus is on the HICO-DET dataset.

\paragraph{Implementation Details}
For HICO-DET and V-COCO datasets, following~\cite{zhang2023pvic, 11092421}, we use the fine-tuned DETR and $\mathcal{H}$-DETR~\cite{Jia2022DETRsWH} models as the object detector. During training, following the commonly used training paradigm of two-stage methods, the weights of the object detector are frozen, while only the interaction predictor is trained. Meanwhile, following~\cite{ting2023hoi, lei2025hola}, we employ the CLIP ViT-B/16 and ViT-L/14@336px models for our method, and all the parameters of them are frozen, except for the adapter layer. We also apply data augmentation techniques from previous works~\cite{ting2023hoi, zhang2023pvic}, including random cropping, resizing, and color jittering. The parameters $\tau$ and $\gamma$ are initialized as 0.1 and constrained between 0 and 1. We empirically set the timesteps $T=2$. The model is optimized using the AdamW optimizer with an initial learning rate of $10^{-4}$ and drops the learning rate by a factor of 5 at the 20th epoch. All models are trained for 30 epochs on 2 NVIDIA A100 GPUs, using a batch size of 16.

\subsection{Comparison with State-of-the-art Methods}
Table~\ref{tab:compare-all} provides a comprehensive performance comparison between our proposed MGNM and current SOTA methods. For clarity, the competing methods are grouped into one-stage and two-stage categories. On the HICO-DET dataset, MGNM outperforms all other one-stage and two-stage approaches that utilize a ResNet backbone. Specifically, under the Default Full setting, our method achieves a 5.02 and 2.93 mAP improvement over the DiffHOI and MP-HOI, despite they leveraging two VLMs, namely CLIP and Stable Diffusion. MGNM also surpasses the leading two-stage approaches, VRDiff and HORP, by 2.66 and 0.82 mAP under the Default Full setting, even though they incorporates additional models. Moreover, MGNM exceeds the performance of the recent GNN-based method, ViPLO, by 4.48 mAP under the Default Full setting, notwithstanding ViPLO's adoption of the supplementary ViTPose-L model~\cite{xu2022vitpose}. In the Known Object setting, MGNM consistently outperforms all ResNet-based counterparts across all three subsets by a substantial margin. Notably, our R50-based MGNM even surpasses FGAHOI-L, which utilizes a more powerful Swin-L backbone~\cite{Liu2021SwinTH}, by 3.54 mAP under the Full setting. Collectively, these results demonstrate the effectiveness of our multimodal graph network and its novel information propagation mechanism. On the V-COCO dataset, MGNM also achieves substantial performance improvement among ResNet-based methods. We note that RLIPv2 obtains a competitive performance on V-COCO, which we attribute to its large-scale pretraining on additional data. This observation suggests that GNN-based approaches, including MGNM, could further benefit from large-scale pretraining, highlighting a promising direction for future research.

To further demonstrate the scalability of our approach, we compare the performance of MGNM equipped with a more advanced object detector against other SOTA methods that utilize similarly strong backbones. As shown in Table~\ref{tab:compare-all}, $\text{MGNM}_{L}$ achieves substantial performance improvements over the competing methods. For example, under the HICO-DET Default Full setting, our method outperforms the Transformer-based methods, PViC-L and Pose-aware, by 5.84 and 4.15 mAP. Notably, although both DiffHOI-L and MP-HOI-L leverage two VLMs each, our single-VLM approach achieves significant gains of 9.53 and 5.63 mAP over these methods, respectively. These results underscore not only the scalability of our framework but also the effectiveness of our multimodal graph network design.

\begin{table}[t]
\begin{center}
\caption{Effectiveness of each component within our framework. w/o means ``without".}
\label{tab:w/o}
\begin{tabular}{c|ccc}
\hline
Methods & Full & Rare & Non-Rare \\
\hline\hline
w/o Spatial Stage & 38.04 & 37.55 & 38.19 \\
w/o Visual Stage & 37.24 & 36.23 & 37.54 \\
w/o Textual Stage & 37.76 & 36.94 & 38.01 \\
w/o Interaction Stage & 37.52 & 35.91 & 38.00 \\
\hline
w/o PDE Diffusion & 37.34 & 35.62 & 37.85 \\
w/o VIS & 37.72 & 36.09 & 38.21 \\
\hline
vanilla & \textbf{39.43} & \textbf{38.69} & \textbf{39.65} \\
\hline
\end{tabular}
\end{center}
\end{table}

\subsection{Ablation Study}
\begin{table}[t]
\begin{center}
\caption{Comparison with different CLIP sizes.}
\label{tab:clip}
\begin{tabular}{l|ccc}
\hline
Methods & Full & Rare & Non-Rare \\
\hline\hline
CLIP-ViT-B/32 & 39.01 & 38.27 & 39.23  \\
CLIP-ViT-B/16 & 39.43 & 38.69 & 39.65 \\
CLIP-ViT-L/14 & 43.41 & 42.59 & 43.66 \\
CLIP-ViT-L/14@336px & 43.94 & 43.15 & 44.17 \\
\hline
\end{tabular}
\end{center}
\end{table}

\paragraph{Components Analysis}
We first analyze the contribution of each of the four stages within our framework, which collectively process low-level and high-level multimodal features. As shown in Table~\ref{tab:w/o}, in row 1 (w/o Spatial Stage), we ablate the spatial features and only use node representations. This results in performance drops of 1.39 mAP. In row 3 (w/o Textual Stage), it leads to a performance drop of 1.67 mAP. This phenomenon demonstrates that these low-level spatial and textual features provide a valuable supplement to the foundational human-object pair representations. Ablating the high-level feature stages, the Visual (row 2) and Interaction (row 4) stages, induces a more substantial performance degradation, with drops of 2.19 and 1.91 mAP, respectively. These decreases are considerably larger than those observed in low-level stages. This finding suggests that the high-level features are more critical to our model's predictive power, suggesting the significance of contextual cues in HOI detection. Next, we analyze the other two important components in our method. In row 5 (w/o PDE Diffusion), we ablate the PDE diffusion mechanism and revert to the conventional MLP fusion mechanism~\cite{zhang2021scg, park2023viplo} to propagate tmultimodal features to the instances. As observed, this modification leads to a significant performance degradation of 2.09 mAP. This demonstrates the effectiveness of our PDE diffusion strategy in modeling the information propagation. In row 6 (w/o VIS), we omit the VIS mechanism to assess its specific contribution. As shown in Table~\ref{tab:w/o}, it results in a performance drop of 1.71 mAP. This highlights the critical role of the VIS mechanism in yielding fine-grained semantic cues for the overall framework. In conclusion, these systematic ablation studies collectively validate the effectiveness of the proposed multimodal graph network framework.

\paragraph{Analysis on CLIP Size}
As a classic vision-language model, CLIP has been widely adopted in HOI detection task and demonstrated to be effective~\cite{ting2023hoi, Lei2024EZHOIVA, park2023viplo}. Since different versions of CLIP differ significantly in model architecture and parameter count, in this subsection, we analyze the influence of different CLIP model scales on our method. As shown in Table~\ref{tab:clip}, we find that CLIP with the same architecture achieves competitive performance. For instance, CLIP-ViT-B/16 outperforms CLIP-ViT-B/32 by 0.42 mAP under the Full setting. This small margin is expected. While ViT-B/16 produces more detailed visual features, they share the same base architecture, and the parameter count is almost same. Meanwhile, other modules in our HOI detector also provide fine-grained features, which likely further mitigates the gap between the ViT-B/32 and ViT-B/16 variants. In contrast, a significant difference emerges between the ViT-B and ViT-L model families. For example, CLIP-ViT-L/14 achieves a significant improvement of 3.98 mAP over CLIP-ViT-B/16. This is an expected improvement, as ViT-L/14 has a larger architecture and possesses approximately $3.5\times$ as many parameters as ViT-B/16, resulting in a much larger model capacity. Additionally, the text encoders in different CLIP sizes are architecturally similar, varying primarily in their projection dimension. Consequently, the improvement is primarily attributable to the more powerful visual encoder. Based on this analysis, adopting a larger CLIP model appears to be an effective strategy for improving performance in HOI detection. We also conduct ablation studies on CLIP sizes with the $\mathcal{H}$-DETR. Although the performance gain is not as significant as that with R50, it still brings an effective improvement of 2.08 mAP. More discussion about this can be found in Appendix B.

\paragraph{Analysis on Timesteps}
\begin{figure}[t]
  \centering
  \includegraphics[width=1.0\linewidth]{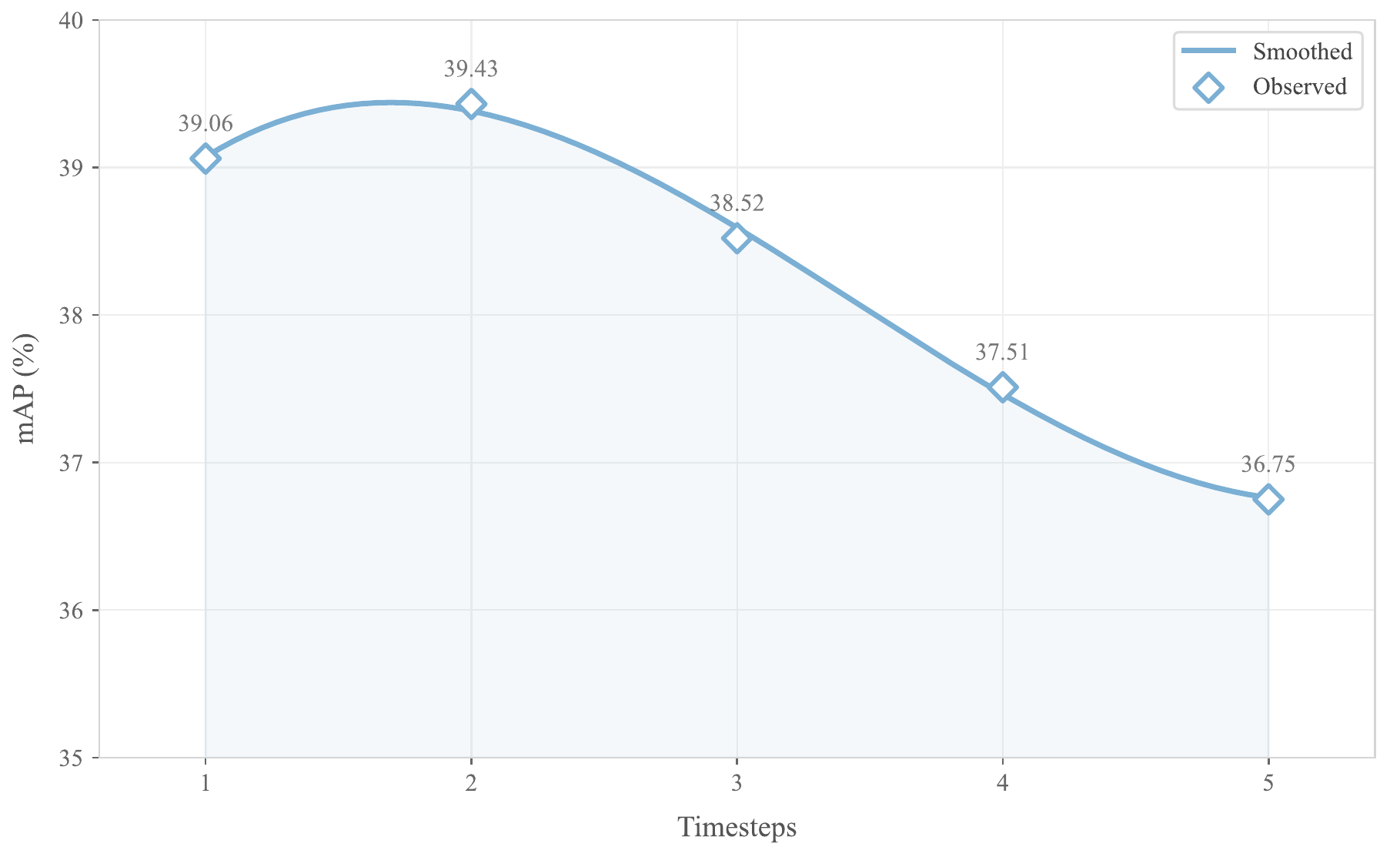}
  \caption{Illustration of performance against timesteps.}
  \label{fig:timesteps}
\end{figure}

As illustrated in Fig.~\ref{fig:timesteps}, our proposed MGNM rapidly achieves optimal performance at $T=2$. Notably, performance gradually degrades as the number of timesteps increases beyond $T=2$. We attribute this degradation to two primary factors: discretization error amplification and gradient attenuation. First, from a numerical computing perspective, the errors from forward-euler method will exponentially accumulate into a severe global discretization error over a long iterative trajectory. This accumulation aggressively distorts the feature manifold and pushes the node representations off the true continuous energy-minimization trajectory. Second, from an optimization perspective, iterating the PDE $T$ times constructs a deep, recurrent computational graph that necessitates backpropagation through time (BPTT). This deep architecture may lead to gradient attenuation during the training process. Furthermore, from a practical standpoint, executing an excessive number of iterations substantially increases inference latency. Therefore, employing a moderate number of timesteps mitigates these numerical and optimization challenges while striking an optimal balance between accuracy and computational efficiency.

\begin{figure*}[t] 
    \centering
    \subfloat[\textit{stand on skateboard}]{%
        \includegraphics[width=0.23\textwidth]{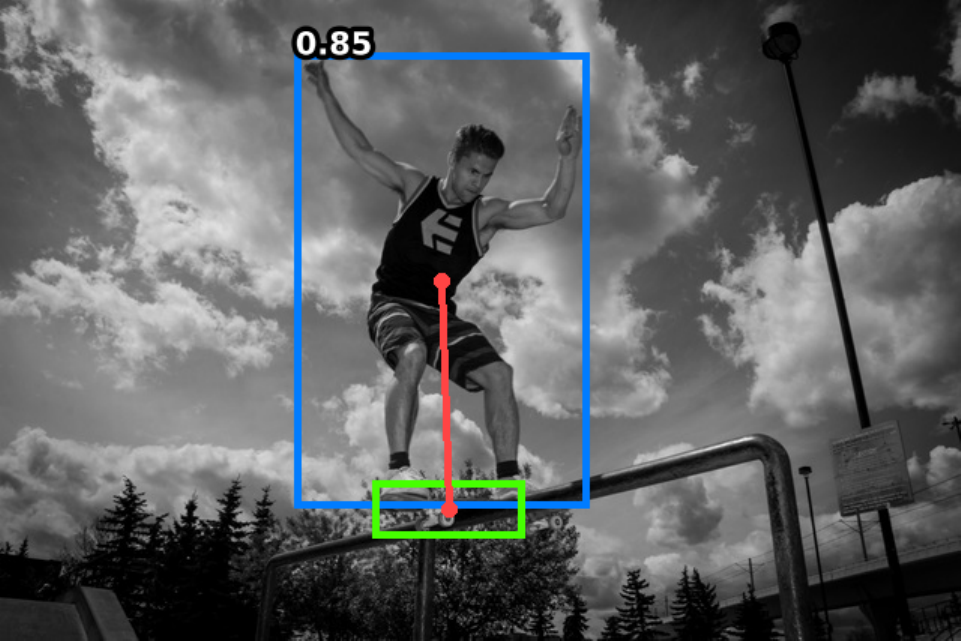}%
        \label{fig:fails:1a}}
    \hfill
    \subfloat[\textit{ride bicycle}]{%
        \includegraphics[width=0.23\textwidth]{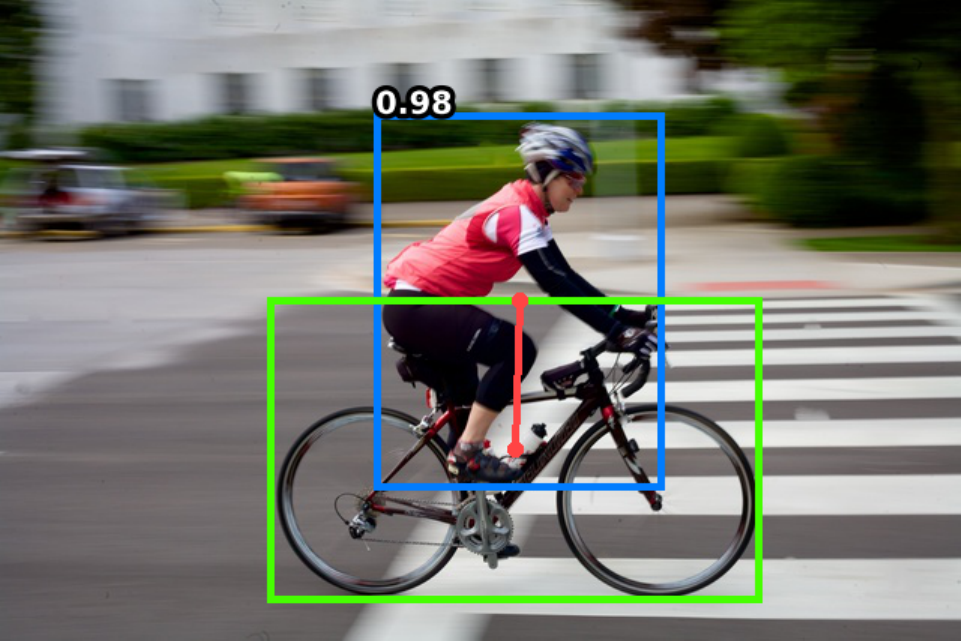}%
        \label{fig:fails:1b}}
    \hfill
    \subfloat[\textit{sit on bench}]{%
        \includegraphics[width=0.23\textwidth]{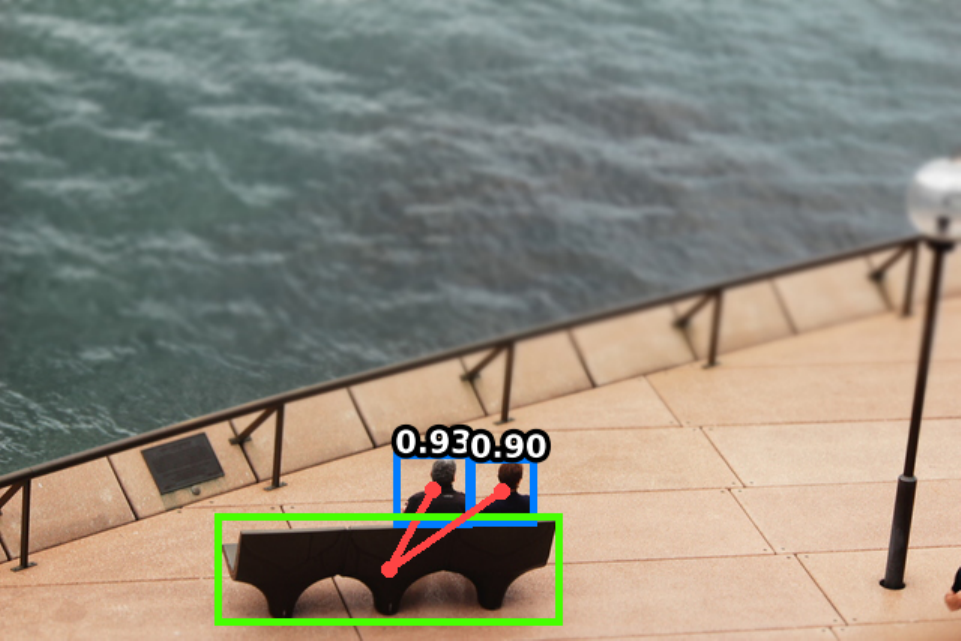}%
        \label{fig:fails:2a}}
    \hfill
    \subfloat[\textit{wear tie}]{%
        \includegraphics[width=0.23\textwidth]{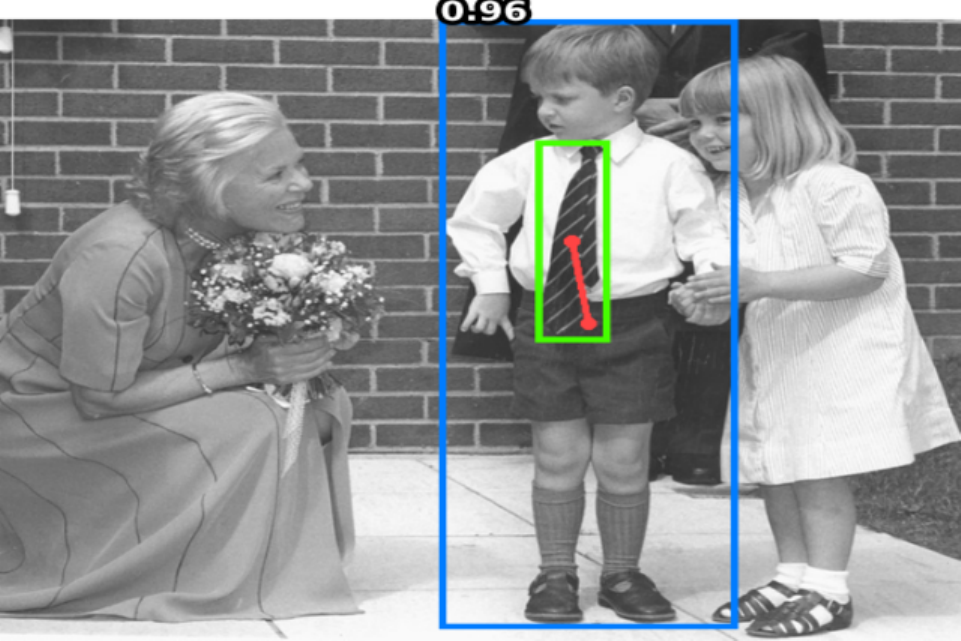}%
        \label{fig:fails:2b}}

    \vspace{1mm} 

    \subfloat[\textit{ride snowboard}]{%
        \includegraphics[width=0.23\textwidth]{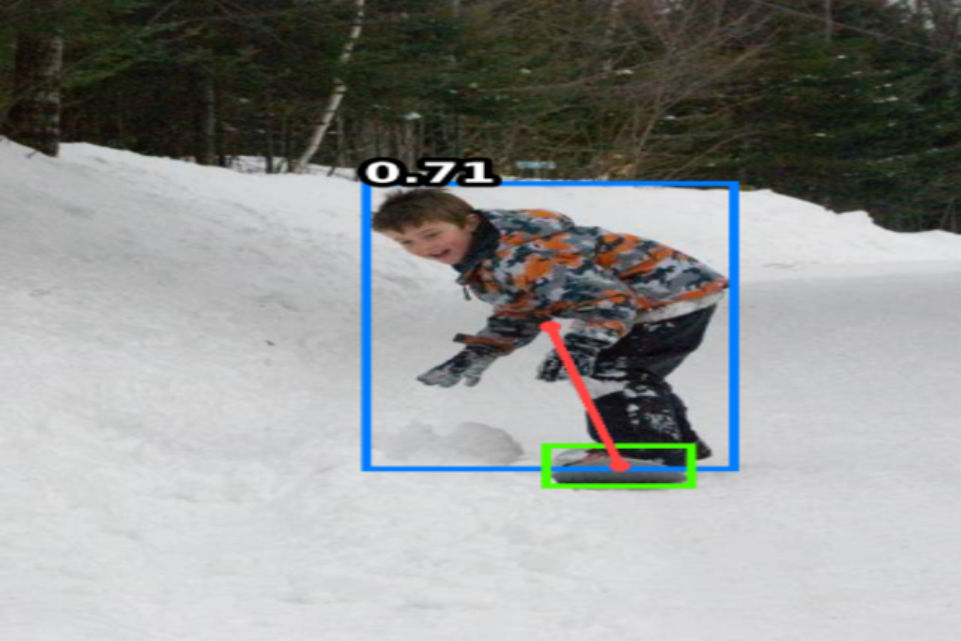}%
        \label{fig:fails:1c}}
    \hfill
    \subfloat[\textit{drive boat}]{%
        \includegraphics[width=0.23\textwidth]{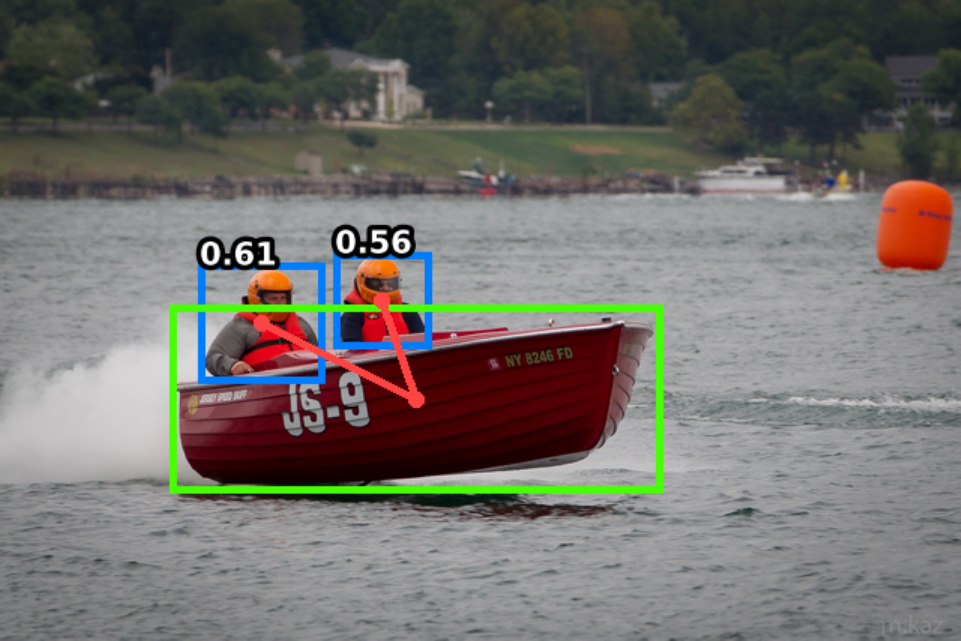}%
        \label{fig:fails:3a}}
    \hfill
    \subfloat[\textit{Visualization for f}]{%
        \includegraphics[width=0.23\textwidth]{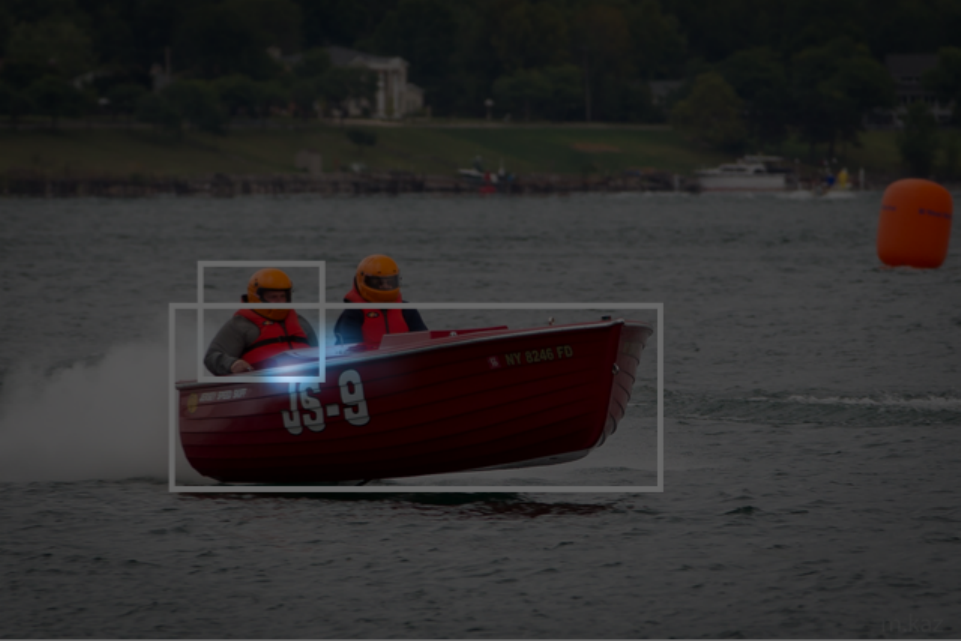}%
        \label{fig:fails:3b}}
    \hfill
    \subfloat[\textit{Visualization for f}]{%
        \includegraphics[width=0.23\textwidth]{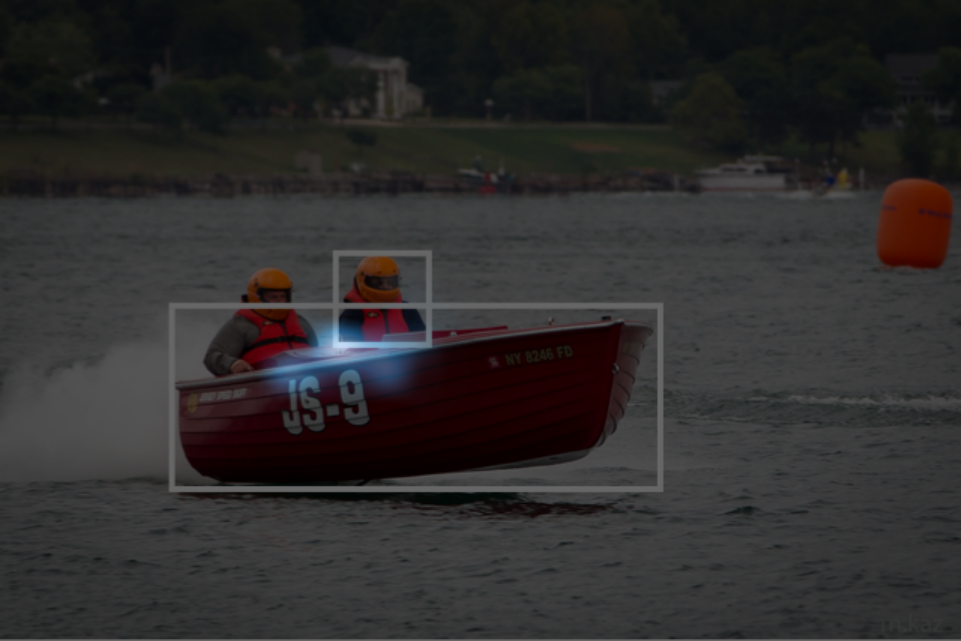}%
        \label{fig:fails:3c}}

    \caption{Qualitative results on the HICO-DET dataset. Fig.~\subref{fig:fails:1a}-\subref{fig:fails:1c} illustrate successful predictions across different challenging scenarios. Fig.~\subref{fig:fails:3a} shows a representative failure case, and Fig.~\subref{fig:fails:3b} and~\subref{fig:fails:3c} visualize its corresponding attention maps.}
    \label{fig:fails}
\end{figure*}

\begin{figure}[t]
  \centering
  \includegraphics[width=1.0\linewidth]{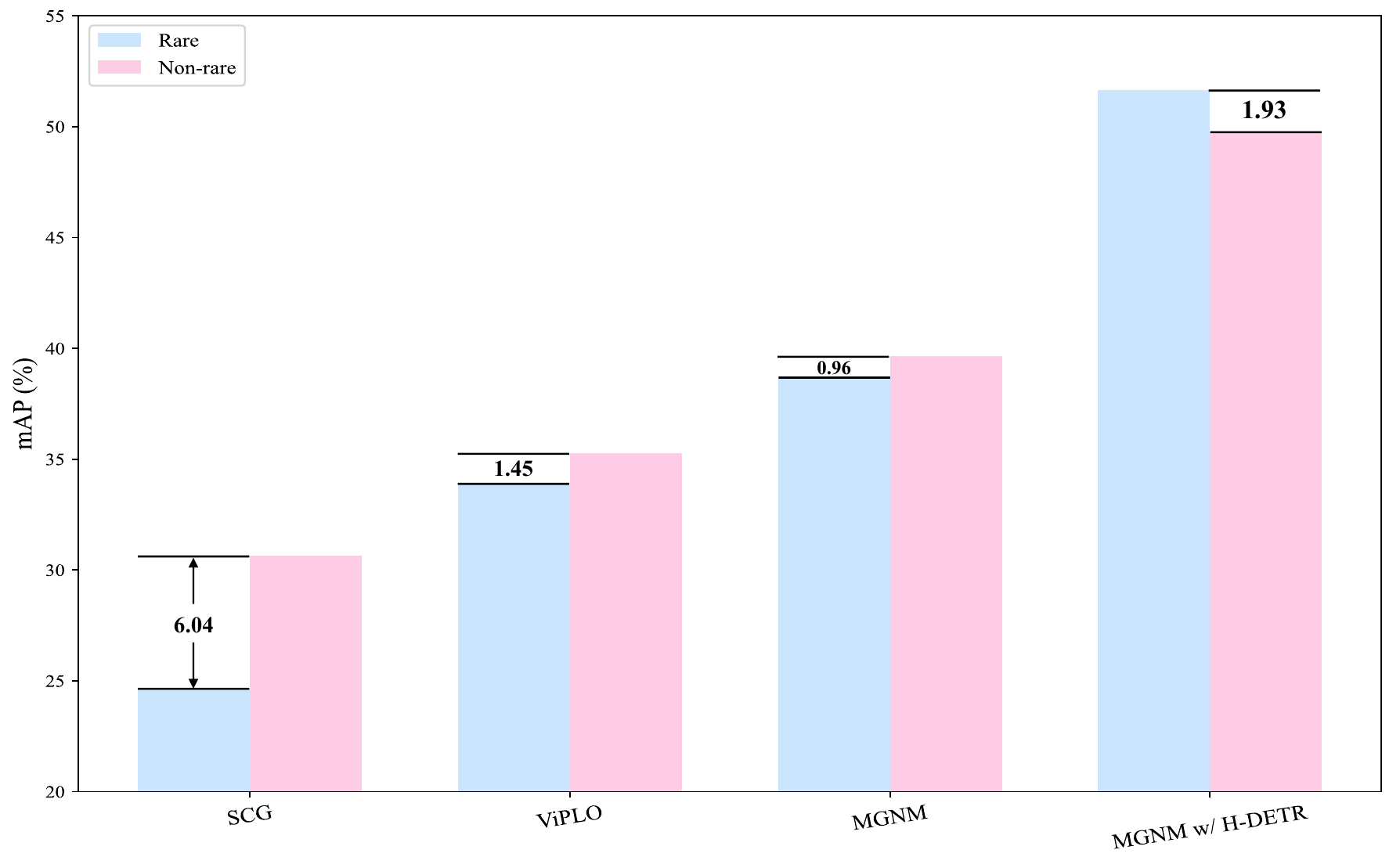}
  \caption{Comparison of our method and two related GNN-based methods on rare and non-rare classes.}
  \label{fig:rare}
\end{figure}

\subsection{Efficiency Analysis}
\begin{table}[t]
\begin{center}
\caption{Efficiency comparison of our proposed MGNM and other methods. TP denotes the trainable parameters and Time denotes the inference time per image. VP means the ViTPose-L model.}
\label{tab:efficiency}
\begin{tabular}{l|c|ccc}
\hline
Methods & Backbone & mAP & Time & TP \\
\hline\hline
SCG~\cite{zhang2021scg} & R50 & 29.26 & 182ms & 54.2M   \\
ViPLO~\cite{park2023viplo} & R101-DC+CLIP+VP & 34.95 & 166ms & 118.2M  \\
HOICLIP~\cite{Ning2023HOICLIPEK} & R50+CLIP & 34.59 & 101ms & 66.1M  \\
CLIP4HOI~\cite{clip4hoi} & R50+CLIP & 35.33 & - & 71.2M  \\
HORP~\cite{11092421} & R50+CLIP+Gaze & 38.61 & 427ms & 26.5M \\
\hline
MGNM w/ MLP & R50+CLIP & 37.34 & 106ms & 20.9M  \\
MGNM & R50+CLIP & \textbf{39.43} & \textbf{72ms} & \textbf{19.5M} \\
\hline
\end{tabular}
\end{center}
\end{table}

To analyze the efficiency of the proposed method, we compare it against existing two-stage methods in terms of mAP, inference time, and trainable parameters. As detailed in Table~\ref{tab:efficiency}, the proposed MGNM consistently achieves significant superiority over other methods. Specifically, when compared to the GNN-based methods SCG and ViPLO, our method achieves $2.5 \times$ and $2.3 \times$ speedups, while utilizing only 36\% and 16\% of their trainable parameters. These results demonstrate the compelling computational efficiency of the proposed PDE graph diffusion mechanism. Furthermore, compared to the SOTA method HORP, which employs HOI-instance-level prompts, our method operates $5.9\times$ faster, with 25\% fewer trainable parameters. This result further highlights the efficiency of our image-level prompt. Moreover, we compare the efficiency of our MGNM with the MLP fusion variant. It can be observed that our MGNM is 34ms faster and requires fewer trainable parameters. These results further demonstrate the efficiency of the overall framework design.

\subsection{Rare-Non-rare Bias}
The challenge of few-shot learning is particularly pronounced in HOI detection, where it manifests as a significant performance disparity between rare and non-rare classes. As evidenced in Table~\ref{tab:compare-all}, this long-tail distribution problem is severe in many methods. For instance, PPDM and FGAHOI exhibit a performance gap of over 10 mAP between their rare and non-rare classes. Similarly, in Fig.~\ref{fig:rare}, the GNN-based method SCG also has a significant difference between rare and non-rare classes. However, a key observation from Table~\ref{tab:compare-all} is that methods incorporating VLMs substantially mitigate this disparity. This suggests that the rich prior knowledge in VLMs can effectively alleviate the Rare-Non-rare bias. Results illustrated in Fig.~\ref{fig:rare} also support this hypothesis: both MGNM and ViPLO, which leverage CLIP, demonstrate a much narrower performance gap between rare and non-rare classes. Moreover, as shown in Table~\ref{tab:compare-all} and Fig.~\ref{fig:rare}, an unexpected trend emerges for methods equipped with a more advanced object detector, such as MP-HOI-L, PViC-L, Pose-aware and our $\text{MGNM}_{L}$. These models exhibit a remarkable reversal, where performance on rare classes actually exceeds that on non-rare classes. We posit that for rare categories, performance bottlenecks stem primarily from the object detection results. Consequently, as detection performance improves, rare categories exhibit a more pronounced performance gain compared to the non-rare categories. In summary, our analysis indicates that leveraging VLMs and advanced object detectors represents a highly effective strategy for mitigating the rare-non-rare bias.

\section{Qualitative Results and Limitations}
\label{sec:qua}
This section presents a qualitative analysis of our method's predictions, highlighting both its strengths and current limitations. As illustrated in Fig.~\ref{fig:fails:1a}-\ref{fig:fails:1c}, our model demonstrates robustness by accurately predicting HOI triplets in single-person or multi-person scenarios, even when the target object is small or occluded. Conversely, Fig.~\ref{fig:fails:3a} presents a representative failure case involving a false positive prediction. Here, the model incorrectly infers that two individuals are driving the boat, a prediction that contradicts common sense in the real world. To diagnose this error, we visualize the corresponding attention maps in Fig.~\ref{fig:fails:3a} and~\ref{fig:fails:3c}. The visualization reveals that the model correctly identifies promising regions associated with the \textit{drive} action for both individuals. However, it lacks common-sense knowledge to understand that only one person can drive a single boat, leading to the erroneous prediction. Furthermore, its attention is diffuse and does not pinpoint fine-grained cues, such as hand-on-wheel placement. This limitation highlights a crucial direction for future work: developing methods with real-world common-sense reasoning to resolve such logical ambiguities.

\section{Conclusion}
\label{sec:con}
In this paper, we introduce multimodal graph network modeling, a novel and effective two-stage HOI detection framework. The central contribution of MGNM lies in its PDE graph diffusion mechanism, which models the information propagation within GNNs from a white-box theoretical perspective. Furthermore, we design a novel variational information squeezing mechanism to further refine the contextual semantics from the multimodal features. Extensive experiments validate the superior effectiveness and efficiency of the proposed method. Additionally, our analysis of the pervasive class imbalance issue provides novel insights and practical strategies for mitigating the rare-non-rare bias in future research.

\bibliographystyle{IEEEtran}
\bibliography{main}

@ARTICLE{11247879,
  author={Wang, Yihan and Sun, Baoli and Li, Haojie and Ma, Xinzhu and Wang, Zhihui and Wang, Zhiyong},
  journal={IEEE TMM}, 
  title={UniAlign: A Universal Cross-Modality Knowledge Alignment Framework for Fine-Grained Action Recognition}, 
  year={2026},
  volume={28},
  number={},
  pages={891-901}
}

@ARTICLE{8848601,
  author={Xu, Bingjie and Li, Junnan and Wong, Yongkang and Zhao, Qi and Kankanhalli, Mohan S.},
  journal={IEEE TMM}, 
  title={Interact as You Intend: Intention-Driven Human-Object Interaction Detection}, 
  year={2020},
  volume={22},
  number={6},
  pages={1423-1432}
}

@inproceedings{wang2024interaction,
  title={Interaction-Centric Spatio-Temporal Context Reasoning for Multi-person Video HOI Recognition},
  author={Wang, Yisong and Xi, Nan and Meng, Jingjing and Yuan, Junsong},
  booktitle={ECCV},
  pages={419--435},
  year={2024}
}

@ARTICLE{10844064,
  author={Zhang, Xiaodan and Jia, Aozhe and Ji, Junzhong and Qu, Liangqiong and Ye, Qixiang},
  journal={IEEE TIP}, 
  title={Intra- and Inter-Head Orthogonal Attention for Image Captioning}, 
  year={2025},
  volume={34},
  number={},
  pages={594-607},
}

@ARTICLE{9489275,
  author={Wang, Haoran and Jiao, Licheng and Liu, Fang and Li, Lingling and Liu, Xu and Ji, Deyi and Gan, Weihao},
  journal={IEEE TIP}, 
  title={IPGN: Interactiveness Proposal Graph Network for Human-Object Interaction Detection}, 
  year={2021},
  volume={30},
  number={},
  pages={6583-6593}
}

@ARTICLE{9410374,
  author={Huang, Qingbao and Liang, Yu and Wei, Jielong and Cai, Yi and Liang, Hanyu and Leung, Ho-fung and Li, Qing},
  journal={IEEE TMM}, 
  title={Image Difference Captioning With Instance-Level Fine-Grained Feature Representation}, 
  year={2022},
  volume={24},
  number={},
  pages={2004-2017}
}

@ARTICLE{10812791,
  author={Gao, Hailiang and Xie, Guo-Sen and Yan, Rui and Cui, Qiongjie and Qu, Hongyu and Shu, Xiangbo},
  journal={IEEE TMM}, 
  title={Hierarchical Motion-Enhanced Matching Framework for Few-Shot Action Recognition}, 
  year={2025},
  volume={27},
  number={},
  pages={2450-2462}
}

@ARTICLE{10097833,
  author={Zhu, Peipei and Wang, Xiao and Zhu, Lin and Sun, Zhenglong and Zheng, Wei-Shi and Wang, Yaowei and Chen, Changwen},
  journal={IEEE TMM}, 
  title={Prompt-Based Learning for Unpaired Image Captioning}, 
  year={2024},
  volume={26},
  number={},
  pages={379-393},
}

@ARTICLE{11194256,
  author={Li, Ming-Zhe and Jia, Zhen and Zhang, Zhang and Li, Yaoning and Ma, Zhanyu and Wang, Liang},
  journal={IEEE TMM}, 
  title={Multi-View Knowledge Guided Semantic Prototype Learning for Generalized Zero-Shot Action Recognition}, 
  year={2025},
  volume={27},
  number={},
  pages={9735-9748},
}

@article{Kingma2013AutoEncodingVB,
  title={Auto-Encoding Variational Bayes},
  author={Diederik P. Kingma and Max Welling},
  journal={ICLR},
  year={2013},
}

@inproceedings{clip4hoi,
author={Mao, Yunyao and Deng, Jiajun and Zhou, Wengang and Li, Li and Fang, Yao and Li, Houqiang},
title={CLIP4HOI: towards adapting CLIP for practical zero-shot HOI detection},
year={2023},
booktitle={NeurIPS},
}

@article{Ning2023HOICLIPEK,
  title={HOICLIP: Efficient Knowledge Transfer for HOI Detection with Vision-Language Models},
  author={Sha Ning and Longtian Qiu and Yongfei Liu and Xuming He},
  journal={CVPR},
  year={2023},
  pages={23507-23517},
}

@ARTICLE{11249431,
  author={Chen, Dongpan and Kong, Dehui and Gao, Junna and Li, Jinghua and Li, Qianxing and Yin, Baocai},
  journal={IEEE TMM}, 
  title={ASK-HOI: Affordance-Scene Knowledge Prompting for Human-Object Interaction Detection}, 
  year={2026},
  volume={28},
  pages={742-756},
}

@INPROCEEDINGS{10658050,
  author={Wang, Guangzhi and Guo, Yangyang and Xu, Ziwei and Kankanhalli, Mohan},
  booktitle={CVPR}, 
  title={Bilateral Adaptation for Human-Object Interaction Detection with Occlusion-Robustness}, 
  year={2024},
  pages={27970-27980},
}

@article{Gkioxari2017DetectingAR,
  title={Detecting and Recognizing Human-Object Interactions},
  author={Georgia Gkioxari and Ross B. Girshick and Piotr Doll{\'a}r and Kaiming He},
  journal={CVPR},
  year={2018},
  pages={8359-8367},
}

@article{Yuan2023RLIPv2FS,
  title={RLIPv2: Fast Scaling of Relational Language-Image Pre-training},
  author={Hangjie Yuan and Shiwei Zhang and Xiang Wang and Samuel Albanie and Yining Pan and Tao Feng and Jianwen Jiang and Dong Ni and Yingya Zhang and Deli Zhao},
  journal={ICCV},
  year={2023},
  pages={21592-21604},
}

@inproceedings{wu2024exploring,
  title={Exploring pose-aware human-object interaction via hybrid learning},
  author={Wu, Eastman ZY and Li, Yali and Wang, Yuan and Wang, Shengjin},
  booktitle={CVPR},
  pages={17815--17825},
  year={2024}
}

@InProceedings{Cao_2025_ICCV,
    author= {Cao, Ping and Tang, Yepeng and Zhang, Chunjie and Zheng, Xiaolong and Liang, Chao and Wei, Yunchao and Zhao, Yao},
    title={Visual Relation Diffusion for Human-Object Interaction Detection},
    booktitle={ICCV},
    year={2025},
    pages={23551-23560}
}

@inproceedings{NEURIPS2024_2a54def4,
 author = {Li, Liulei and Wang, Wenguan and Yang, Yi},
 booktitle = {NeurIPS},
 pages = {23655--23678},
 title = {Human-Object Interaction Detection Collaborated with Large Relation-driven Diffusion Models},
 year = {2024}
}

@inproceedings{Vaswani2017AttentionIA,
  title={Attention is All you Need},
  author={Ashish Vaswani and Noam M. Shazeer and Niki Parmar and Jakob Uszkoreit and Llion Jones and Aidan N. Gomez and Lukasz Kaiser and Illia Polosukhin},
  booktitle={NeurIPS},
  year={2017},
}

@inproceedings{Qi2018LearningHI,
  title={Learning Human-Object Interactions by Graph Parsing Neural Networks},
  author={Siyuan Qi and Wenguan Wang and Baoxiong Jia and Jianbing Shen and Song-Chun Zhu},
  booktitle={ECCV},
  year={2018},
}

@inproceedings{Gao2020DRGDR,
  title={DRG: Dual Relation Graph for Human-Object Interaction Detection},
  author={Chen Gao and Jiarui Xu and Yuliang Zou and Jia-Bin Huang},
  booktitle={ECCV},
  year={2020},
}

@article{Zhong2021GlanceAG,
  title={Glance and Gaze: Inferring Action-aware Points for One-Stage Human-Object Interaction Detection},
  author={Xubin Zhong and Xian Qu and Changxing Ding and Dacheng Tao},
  journal={CVPR},
  year={2021},
  pages={13229-13238},
}

@inproceedings{Jia2024ContextHOISC,
  title={ContextHOI: Spatial Context Learning for Human-Object Interaction Detection},
  author={Mingda Jia and Liming Zhao and Ge Li and Yun Zheng},
  booktitle={AAAI},
  year={2025}
}

@article{Xue2025GuidingHI,
  title={Guiding Human-Object Interactions with Rich Geometry and Relations},
  author={Mengqing Xue and Yifei Liu and Ling Guo and Shaoli Huang and Changxing Ding},
  journal={CVPR},
  year={2025},
  pages={22714-22723},
}

@inproceedings{Jia2024OrchestratingTS,
  title={Orchestrating the Symphony of Prompt Distribution Learning for Human-Object Interaction Detection},
  author={Mingda Jia and Liming Zhao and Ge Li and Yun Zheng},
  booktitle={AAAI},
  year={2024},
}

@ARTICLE{10227593,
  author={Fang, Shuman and Lin, Zhiwen and Yan, Ke and Li, Jie and Lin, Xianming and Ji, Rongrong},
  journal={IEEE TMM}, 
  title={HODN: Disentangling Human-Object Feature for HOI Detection}, 
  year={2024},
  volume={26},
  pages={3125-3136},
}

@article{Yang2025NoMS,
  title={No More Sibling Rivalry: Debiasing Human-Object Interaction Detection},
  author={Bin Yang and Yulin Zhang and Hong-Yu Zhou and Sibei Yang},
  journal={ICCV},
  year={2025},
  volume={abs/2509.00760},
}

@INPROCEEDINGS{11092421,
  author={Geng, Pei and Yang, Jian and Zhang, Shanshan},
  booktitle={CVPR}, 
  title={HORP: Human-Object Relation Priors Guided HOI Detection}, 
  year={2025},
  pages={25325-25335},
}

@inproceedings{lei2025hola,
title={HOLa: Zero-Shot HOI Detection with Low-Rank Decomposed VLM Feature Adaptation},
author={Lei, Qinqian and Wang, Bo and Robby T., Tan},
booktitle={ICCV},
year={2025}
}

@article{Jia2022DETRsWH,
  title={DETRs with Hybrid Matching},
  author={Ding Jia and Yuhui Yuan and Hao He and Xiao-pei Wu and Haojun Yu and Weihong Lin and Lei-huan Sun and Chao Zhang and Hanhua Hu},
  journal={CVPR},
  year={2023},
  pages={19702-19712},
}

@article{Liu2021SwinTH,
  title={Swin Transformer: Hierarchical Vision Transformer using Shifted Windows},
  author={Ze Liu and Yutong Lin and Yue Cao and Han Hu and Yixuan Wei and Zheng Zhang and Stephen Lin and Baining Guo},
  journal={ICCV},
  year={2021},
  pages={9992-10002},
}

@INPROCEEDINGS{10889833,
  author={Nan, Fang and Zhang, Ni and Liu, Qidong and Jing, Wei and Dai, Guang and Chen, Yan and Tian, Feng},
  booktitle={ICASSP}, 
  title={Exploring Triple Knowledge Cues for Zero-Shot Human-Object Interaction Detection}, 
  year={2025},
  pages={1-5},
}

@article{lu2025intra,
  title={Intra-and inter-instance location correlation network for human--object interaction detection},
  author={Lu, Minglang and Yang, Guanci and Wang, Yang and Luo, Kexin},
  journal={Eng. Appl. Artif. Intell.},
  volume={142},
  pages={109942},
  year={2025},
}

@ARTICLE{10242152,
  author={Zong, Daoming and Sun, Shiliang},
  journal={IEEE TNNLS}, 
  title={Zero-Shot Human–Object Interaction Detection via Similarity Propagation}, 
  year={2024},
  volume={35},
  number={12},
  pages={17805-17816}
}

@inproceedings{kim2025locality,
  title={Locality-Aware Zero-Shot Human-Object Interaction Detection},
  author={Kim, Sanghyun and Jung, Deunsol and Cho, Minsu},
  booktitle={CVPR},
  pages={20190--20200},
  year={2025}
}

@article{Rombach2021HighResolutionIS,
  title={High-Resolution Image Synthesis with Latent Diffusion Models},
  author={Robin Rombach and A. Blattmann and Dominik Lorenz and Patrick Esser and Bj{\"o}rn Ommer},
  journal={CVPR},
  year={2021},
  pages={10674-10685}
}

@article{Yang2024OpenWorldHI,
  title={Open-World Human-Object Interaction Detection via Multi-Modal Prompts},
  author={Jie Yang and Bingliang Li and Ailing Zeng and Lei Zhang and Ruimao Zhang},
  journal={CVPR},
  year={2024},
  pages={16954-16964},
}

@inproceedings{qu2022distillation,
  title={Distillation using oracle queries for transformer-based human-object interaction detection},
  author={Qu, Xian and Ding, Changxing and Li, Xingao and Zhong, Xubin and Tao, Dacheng},
  booktitle={CVPR},
  pages={19558--19567},
  year={2022}
}

@ARTICLE{10496247,
  author={Liao, Yue and Liu, Si and Gao, Yulu and Zhang, Aixi and Li, Zhimin and Wang, Fei and Li, Bo},
  journal={IEEE TPAMI}, 
  title={PPDM++: Parallel Point Detection and Matching for Fast and Accurate HOI Detection}, 
  year={2024},
  volume={46},
  number={10},
  pages={6826-6841},
}

@ARTICLE{10315071,
  author={Ma, Shuailei and Wang, Yuefeng and Wang, Shanze and Wei, Ying},
  journal={IEEE TPAMI}, 
  title={FGAHOI: Fine-Grained Anchors for Human-Object Interaction Detection}, 
  year={2024},
  volume={46},
  number={4},
  pages={2415-2429},
}

@article{Lin2017FocalLF,
  title={Focal Loss for Dense Object Detection},
  author={Tsung-Yi Lin and Priya Goyal and Ross B. Girshick and Kaiming He and Piotr Doll{\'a}r},
  journal={IEEE TPAMI},
  year={2017},
  volume={42},
  pages={318-327},
}

@inproceedings{xu2022vitpose,
  title={Vi{TP}ose: Simple Vision Transformer Baselines for Human Pose Estimation},
  author={Yufei Xu and Jing Zhang and Qiming Zhang and Dacheng Tao},
  booktitle={NeurIPS},
  year={2022},
}

@inproceedings{lei2024efficient,
title={EZ-HOI: VLM Adaptation via Guided Prompt Learning for Zero-Shot HOI Detection},
author={Lei, Qinqian and Wang, Bo and Robby T., Tan},
booktitle={NeurIPS},
year={2024}
}

@inproceedings{ting2023hoi,
  title={Efficient Adaptive Human-Object Interaction Detection with Concept-guided Memory},
  author={Ting Lei and Fabian Caba and Qingchao Chen and Hailin Ji and Yuxin Peng and Yang Liu},
  year={2023},
  booktitle={ICCV},
}

@article{yang2023boosting,
    title={Boosting Human-Object Interaction Detection with Text-to-Image Diffusion Model},
    author={Yang, Jie and Li, Bingliang and Yang, Fengyu and Zeng, Ailing and Zhang, Lei and Zhang, Ruimao},
    journal={ArXiv},
    year={2023}
}

@inproceedings{chen_2021_asnet,
  author = {Chen, Mingfei and Liao, Yue and Liu, Si and Chen, Zhiyuan and Wang, Fei and Qian, Chen},
  title = {Reformulating HOI Detection as Adaptive Set Prediction},
  booktitle={CVPR},
  year = {2021},
}

@article{gupta2015visual,
  title={Visual semantic role labeling},
  author={Gupta, Saurabh and Malik, Jitendra},
  journal={ArXiv},
  year={2015}
}

@inproceedings{chao2018learning,
  title={Learning to detect human-object interactions},
  author={Chao, Yu-Wei and Liu, Yunfan and Liu, Xieyang and Zeng, Huayi and Deng, Jia},
  booktitle={WACV},
  pages={381--389},
  year={2018},
}

@inproceedings{Radford2021LearningTV,
  title={Learning Transferable Visual Models From Natural Language Supervision},
  author={Alec Radford and Jong Wook Kim and Chris Hallacy and Aditya Ramesh and Gabriel Goh and Sandhini Agarwal and Girish Sastry and Amanda Askell and Pamela Mishkin and Jack Clark and Gretchen Krueger and Ilya Sutskever},
  booktitle={ICML},
  year={2021},
}

@article{Guo2024UnseenNM,
  title={Unseen No More: Unlocking the Potential of CLIP for Generative Zero-shot HOI Detection},
  author={Yixin Guo and Yu Liu and Jianghao Li and Weimin Wang and Qi Jia},
  journal={ACM MM},
  year={2024},
}

@article{Lei2024EZHOIVA,
  title={EZ-HOI: VLM Adaptation via Guided Prompt Learning for Zero-Shot HOI Detection},
  author={Qinqian Lei and Bo Wang and Robby T. Tan},
  journal={NeurIPS},
  year={2024},
}

@article{park2023viplo,
      title={ViPLO: Vision Transformer based Pose-Conditioned Self-Loop Graph for Human-Object Interaction Detection}, 
      author={Jeeseung Park and Jin-Woo Park and Jong-Seok Lee},
      journal={CVPR},
      year={2023},
}

@inproceedings{zhang2023pvic,
  author    = {Zhang, Frederic Z. and Yuan, Yuhui and Campbell, Dylan and Zhong, Zhuoyao and Gould, Stephen},
  title     = {Exploring Predicate Visual Context in Detecting Human–Object Interactions},
  booktitle = {ICCV},
  year      = {2023},
  pages     = {10411-10421},
}

@inproceedings{zhang2022upt,
  author    = {Zhang, Frederic Z. and Campbell, Dylan and Gould, Stephen},
  title     = {Efficient Two-Stage Detection of Human-Object Interactions with a Novel Unary-Pairwise Transformer},
  booktitle = {CVPR},
  year      = {2022},
  pages     = {20104-20112}
}

@inproceedings{zhang2021scg,
  author    = {Zhang, Frederic Z. and Campbell, Dylan and Gould, Stephen},
  title     = {Spatially Conditioned Graphs for Detecting Human–Object Interactions},
  booktitle = {ICCV},
  year      = {2021},
  pages     = {13319-13327}
}

@article{Carion2020EndtoEndOD,
  title={End-to-End Object Detection with Transformers},
  author={Nicolas Carion and Francisco Massa and Gabriel Synnaeve and Nicolas Usunier and Alexander Kirillov and Sergey Zagoruyko},
  journal={ECCV},
  year={2020},
}

@article{Liao2022GENVLKTSA,
  title={GEN-VLKT: Simplify Association and Enhance Interaction Understanding for HOI Detection},
  author={Yue Liao and Aixi Zhang and Miao Lu and Yongliang Wang and Xiaobo Li and Si Liu},
  journal={CVPR},
  year={2022},
  pages={20091-20100},
}

@article{Zhou2022HumanObjectID,
  title={Human-Object Interaction Detection via Disentangled Transformer},
  author={Desen Zhou and Zhichao Liu and Jian Wang and Leshan Wang and T. Hu and Errui Ding and Jingdong Wang},
  journal={CVPR},
  year={2022},
  pages={19546-19555},
}

@inproceedings{kim2020uniondet,
  title={Uniondet: Union-level detector towards real-time human-object interaction detection},
  author={Kim, Bumsoo and Choi, Taeho and Kang, Jaewoo and Kim, Hyunwoo J},
  booktitle={ECCV},
  pages={498--514},
  year={2020},
}

@inproceedings{liao2020ppdm,
  title={Ppdm: Parallel point detection and matching for real-time human-object interaction detection},
  author={Liao, Yue and Liu, Si and Wang, Fei and Chen, Yanjie and Qian, Chen and Feng, Jiashi},
  booktitle={CVPR},
  pages={482--490},
  year={2020}
}

@ARTICLE{11264347,
  author={Zhang, Ruonan and Hao, Yiqing and Zhang, Feng and An, Gaoyun and Song, Binyang and Wu, Dapeng Oliver},
  journal={IEEE TPAMI}, 
  title={Human-Inspired Scene Understanding: A Grounded Cognition Method for Unbiased Scene Graph Generation}, 
  year={2026},
  volume={48},
  number={3},
  pages={3286-3303},
}

@article{ma2024scene,
  title={Scene Understanding Method Utilizing Global Visual and Spatial Interaction Features for Safety Production},
  author={Ma, Fuqi and Wang, Bo and Dong, Xuzhu and Li, Min and Ma, Hengrui and Jia, Rong and Jain, Amar},
  journal={Inf. Fusion},
  pages={102668},
  year={2024},
}

\end{document}